\documentclass[journal,twoside,web]{ieeecolor}
\usepackage{generic}
\usepackage{cite}
\usepackage{amsmath,amssymb,amsfonts}
\usepackage{algorithmic}
\usepackage{graphicx}
\usepackage{algorithm,algorithmic}
\usepackage{hyperref}

\usepackage{booktabs}
\usepackage{multirow}
\usepackage{float}

\hypersetup{hidelinks=true}
\usepackage{textcomp}
\def\BibTeX{{\rm B\kern-.05em{\sc i\kern-.025em b}\kern-.08em
    T\kern-.1667em\lower.7ex\hbox{E}\kern-.125emX}}
\begin{document}
\title{SurgWorld: Learning Surgical Representations Beyond Observation via Latent Prediction}
\author{Jingkun Yue, Pengxiang Wang, Siqi Zhang, Zinan Jia, Yiyao Zhang, Huihuan Xu, \\Yichang Hao, Xiaohong Liu, Zongbo Han, Guangyu Wang
\thanks{
This study was supported by the National Natural Science Foundation of China (T2522008 and 62501406 to G.W.; 624B2100 to Z.H.), the National Key R\&D Program of China (2025YFF0515300, 2025YFF0515301, and 2024YFF1207103), and the Shenzhen Medical Research Fund (E250200622 to G.W.). Additional support was provided by the Scientific Research Innovation Capability Support Project for Young Faculty (SRICSPYF-ZY2025015 to G.W.), and the Fundamental Research Funds for Beijing University of Posts and Telecommunications (Grant 2025AI4S18 to G.W.). This work was also supported by the New Cornerstone Science Foundation through the XPLORER PRIZE, and funded by the Beijing University of Posts and Telecommunications-China Mobile Communications Group Co., Ltd. Joint Institute. The funders had no role in study design, data collection and analysis, decision to publish, or preparation of the manuscript.
}
\thanks{
Jingkun Yue and Pengxiang Wang are co-first authors.
(Corresponding author: Guangyu Wang.)
}
\thanks{
Jingkun Yue, Pengxiang Wang, Siqi Zhang, Zinan Jia, Huihuan Xu, Zongbo Han and Guangyu Wang are with the State Key Laboratory of Networking and Switching Technology, Beijing University of Posts and Telecommunications, Beijing, China (yuejk@bupt.edu.cn; guangyu.wang24@gmail.com).}
\thanks{Yichang Hao is with Department of Urology, Peking University Third Hospital.}
\thanks{Yiyao Zhang and Xiaohong Liu are with Institute of Medical Artificial Intelligence, South China Hospital, Medical School, Shenzhen University, Shenzhen, Guangdong, China.}
}

\maketitle

\begin{abstract}
Surgical procedures are inherently partially observable, requiring surgeons to continuously infer evolving anatomical dynamics and future surgical states beyond the current observation. However, existing self-supervised learning methods primarily learn from observed surgical videos through masked reconstruction or predicting representations within the camera's current field of view, lacking the capability to explicitly model the out-of-sight surgical dynamics. 
Inspired by latent predictive world models, we introduce SurgWorld, a video-native, self-supervised framework for surgical representation learning, which treats a specified spatial or temporal shift as a pseudo-action and learns to predict the target latent representation from a partial observation.
Specifically, we propose Out-of-Sight Dynamics Modeling (OSDM) as the pretraining objective for predicting representations conditioned on relative spatio-temporal shifts. 
We further regularize the learned latent dynamics from OSDM with a Geometric Visual Similarity (GVS) objective by aligning patch-level relational structures between predicted and target representations, preserving spatial geometry and local visual coherence for dense perception.
These two objectives are integrated through an agglomerative training framework that consolidates complementary supervisory signals from multiple teachers to learn a unified latent representation of surgical dynamics and visual geometry.
Extensive transfer learning experiments on 17 surgical benchmarks show that SurgWorld consistently outperforms existing vision foundation models across surgical workflow recognition, fine-grained action understanding, and dense visual perception.
We further conduct tailored qualitative and quantitative analyses, demonstrating that SurgWorld effectively captures surgical dynamics beyond the directly observed context through latent prediction.
\end{abstract}


\section{Introduction}
\label{sec:introduction}
\IEEEPARstart{S}{urgical} procedures require a continuous understanding of complex anatomical structures and dynamic instrument-tissue interactions \cite{kiyasseh2023vision}. However, surgeons can only observe a limited portion of the operative field at any given time due to camera viewpoints, occlusions, and the three-dimensional nature of surgical environments \cite{hsieh2026extrags}. Consequently, they must continuously infer latent anatomical structures and anticipate future surgical states from partial observations. The ability to model these out-of-sight surgical dynamics beyond current observations is therefore fundamental for surgical understanding, navigation, and intelligent assistance.

Recent advances in self-supervised learning have enabled surgical foundation models to learn transferable representations from large-scale unlabeled surgical videos, substantially improving performance across diverse downstream tasks \cite{yang2026large}. Existing methods mainly follow two paradigms: reconstruction-based learning (e.g., MAE \cite{he2022masked} and VideoMAE \cite{tong2022videomae}), which recovers masked pixels or video patches, and joint embedding predictive learning borrowed from world-modeling approaches (e.g., V-JEPA \cite{assran2025v, mur2026v}), which predict target representations in the latent space rather than reconstructing raw observations.

Despite their distinct objectives, both paradigms ultimately learn exclusively from observed surgical videos. Reconstruction-based methods (e.g., EndoViT \cite{batic2024endovit}) recover missing content within visible observations, while JEPA-based methods (e.g., SurgMotion \cite{wu2026unisurg}) predict representations of future observations that remain within the camera's field of view. Neither explicitly models out-of-sight surgical dynamics, namely the evolution of hidden anatomical structures and instrument-tissue interactions beyond current observations across space and time. 

\begin{figure}[t]
    \centering
    \includegraphics[width=\columnwidth]{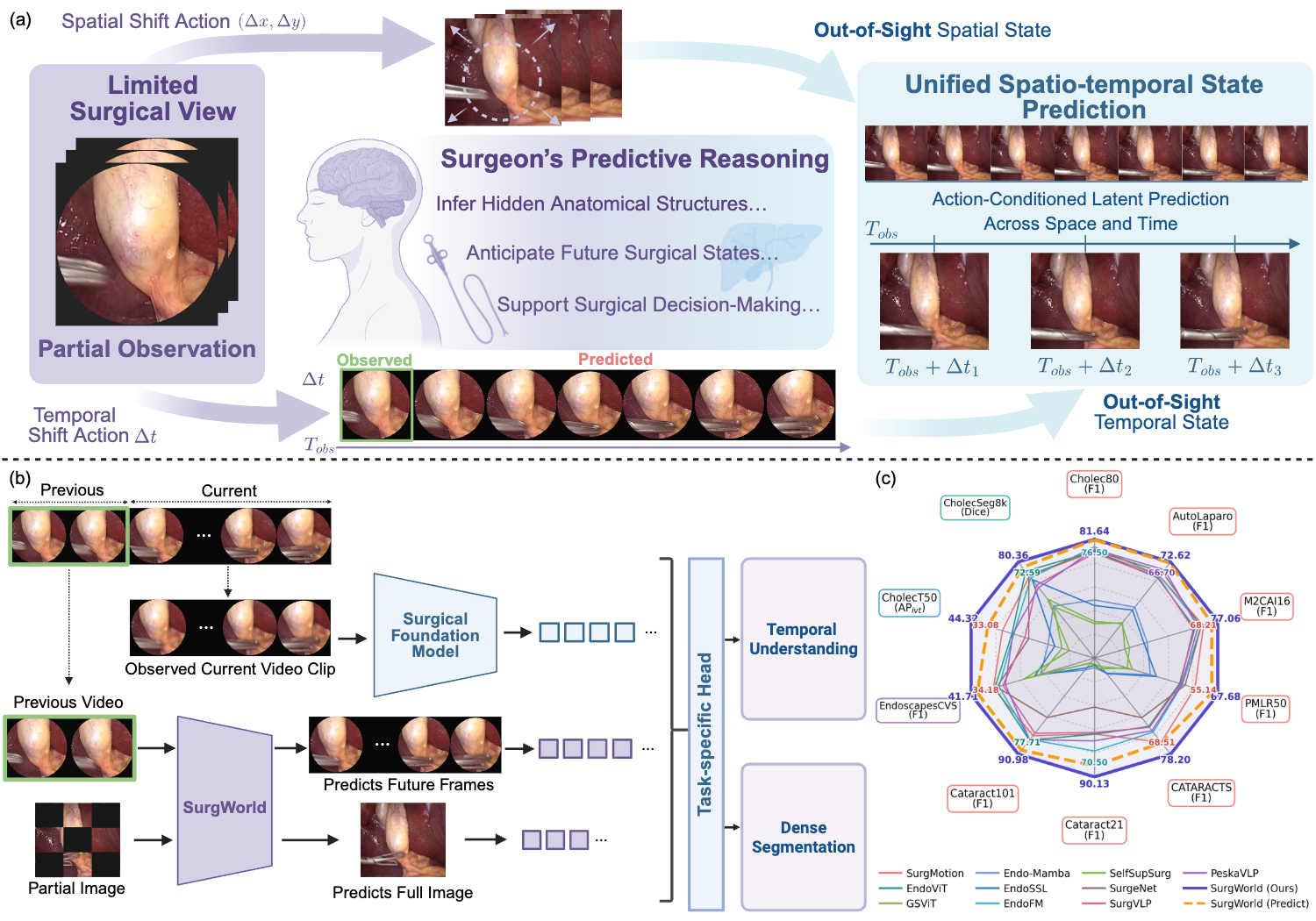}
    \caption{(a) Motivation: From partial surgical observations, surgeons leverage anatomical knowledge, procedural experience, and mental simulation to infer out-of-sight surgical states across space and time, supporting surgical understanding and decision-making. (b) SurgWorld first predicts unobserved temporal or spatial surgical states from partial observations, and then constructs complete representations for temporal understanding and segmentation, whereas conventional surgical foundation models rely on fully observed inputs. (c) SurgWorld achieves state-of-the-art performance across all evaluated surgical benchmarks.}
    \label{fig:fig1}
\end{figure}

To bridge this gap, inspired by latent predictive world models, we introduce SurgWorld, a video-native, self-supervised framework for surgical representation learning, which explicitly predicts the out-of-sight surgical state at specified spatial location or temporal moment.
Based on our observation illustrated in Fig.~\ref{fig:fig1}(a), we propose Out-of-Sight Dynamics Modeling (OSDM), a world-model-inspired pretraining task that formulates surgical video pretraining as action-conditioned latent representation prediction. Given context and target clips sampled from different spatial regions or temporal moments, OSDM represents their relative spatio-temporal shift as a pseudo-action and learns to predict the target latent representation from the observed context under the supervision of an exponential moving average teacher. By learning latent prediction conditioned on relative spatio-temporal shifts, OSDM enables predictive reasoning over out-of-sight surgical states. 
To regularize the latent dynamics learned by OSDM, we further introduce a Geometric Visual Similarity (GVS) objective that aligns patch-wise visual similarity structures with those produced by a frozen dense-feature teacher. This geometry-aware supervision preserves local visual coherence and spatial correspondence, resulting in higher-quality dense representations of surgical scenes.
Finally, the two objectives are integrated within an agglomerative training framework that consolidates complementary supervisory signals from multiple teachers to learn a unified latent representation of surgical dynamics and visual geometry. 

We extensively evaluate SurgWorld on 17 surgical benchmarks spanning surgical workflow recognition, fine-grained action understanding, and dense visual perception. SurgWorld consistently outperforms a diverse range of general-purpose and surgical foundation models. Moreover,  we validate that SurgWorld effectively captures and predicts out-of-sight surgical dynamics through a series of analytical experiments, including future latent prediction under incomplete observations, pseudo-action sensitivity analysis through action-conditioned target-state retrieval, and generative visualization of predicted latent representations for qualitative interpretation.

Our contributions are as follows:

\begin{itemize}
    \item We introduce SurgWorld, a video-native, world-model-inspired self-supervised framework for surgical representation learning that extends beyond visible observations by predicting out-of-sight surgical states. Notably, SurgWorld is trained on 348 hours of publicly available surgical videos without relying on private surgical data, supporting transparency and reproducibility.
    \item We propose Out-of-Sight Dynamics Modeling (OSDM), which treats relative spatio-temporal shifts as pseudo-actions for out-of-sight latent state prediction, regularized by a Geometric Visual Similarity (GVS) objective that preserves spatial geometry and local visual coherence.
    \item SurgWorld achieves state-of-the-art (SOTA) performance on 17 widely used surgical benchmarks. Qualitative and quantitative analyses further demonstrate SurgWorld's ability to predict out-of-sight surgical states under partial observability.

\end{itemize}

\section{Related Work}

\subsection{Self-supervised Learning in Surgery}
Self-supervised learning has become an effective approach to reducing reliance on expert annotations for surgical video understanding. Early studies primarily adapted image-level self-supervised methods by treating videos as collections of independent frames \cite{batic2024endovit, hirsch2023self, schmidgall2024general}, thereby overlooking the continuous evolution of anatomy, instrument motion, tissue deformation, and surgical actions. Recent video-native surgical foundation models have begun to capture these temporal dynamics, demonstrating their importance for transferable surgical representations \cite{yang2026large, wu2026unisurg}.
Existing methods primarily encode or reconstruct observed content, without explicitly modeling the out-of-sight surgical dynamics. Their reliance on proprietary or partially released pre-training data further limits reproducibility and fair comparison. SurgWorld instead formulates surgical representation learning as action-conditioned latent representation prediction, predicting out-of-sight latent states under specified spatial or temporal shifts. Trained exclusively on public datasets, it provides a transparent and reproducible framework for surgical video understanding.

\subsection{World Model}

World models are rooted in cognitive theories that attribute human intelligence to internal representations of latent world states \cite{lecun2022path}. 
Generative world models learn environmental dynamics from large-scale data and predict future observations conditioned on current states and candidate actions, enabling agents to evaluate actions through imagined rollouts rather than costly real-world interactions \cite{bar2025navigation}. This paradigm has been widely explored in model-based reinforcement learning and embodied AI \cite{hafner2025training}, and has recently been applied to controllable surgical video generation for robotic surgery \cite{he2025surgworld}.
A complementary line of world models learns predictive latent representations of the environment, more closely resembling the internal models hypothesized in human cognition. This paradigm has recently been extended to medical imaging~\cite{wang2026towards,noori2026world}: CheXWorld \cite{yue2025chexworld} models anatomical structures and domain variations in radiographs, X-WIN \cite{yang2026x} transfers 3D spatial knowledge from CT volumes through latent projection prediction, and EchoWorld \cite{yue2025echoworld} captures probe-conditioned visual dynamics in echocardiography. Borrowing from this paradigm, SurgWorld adopts a latent predictive formulation for surgical video representation learning, using relative spatio-temporal displacement as a pseudo-action to condition the prediction of target latent representations beyond the directly observed context, thereby modeling the spatial geometry and temporal dynamics of the surgical environment.

\subsection{Agglomerative Vision Foundation Models}
Agglomerative vision models, introduced by AM-RADIO \cite{ranzinger2024radio}, build vision foundation models via multi-teacher distillation by distilling heterogeneous representations from multiple pre-trained vision foundation models into a single student backbone. 
This paradigm offers a promising and empirically validated path toward unified visual representations. 
Inspired by these works \cite{wang2024sam}, we incorporate a teacher model with strong dense feature coherence into the training of SurgWorld. Together with our proposed geometric visual similarity loss, this teacher provides spatially structured supervision that encourages SurgWorld to preserve local visual consistency and geometric correspondence, thereby improving its performance on dense perception.

\begin{figure*}[t]
    \centering
    \includegraphics[width=\linewidth]{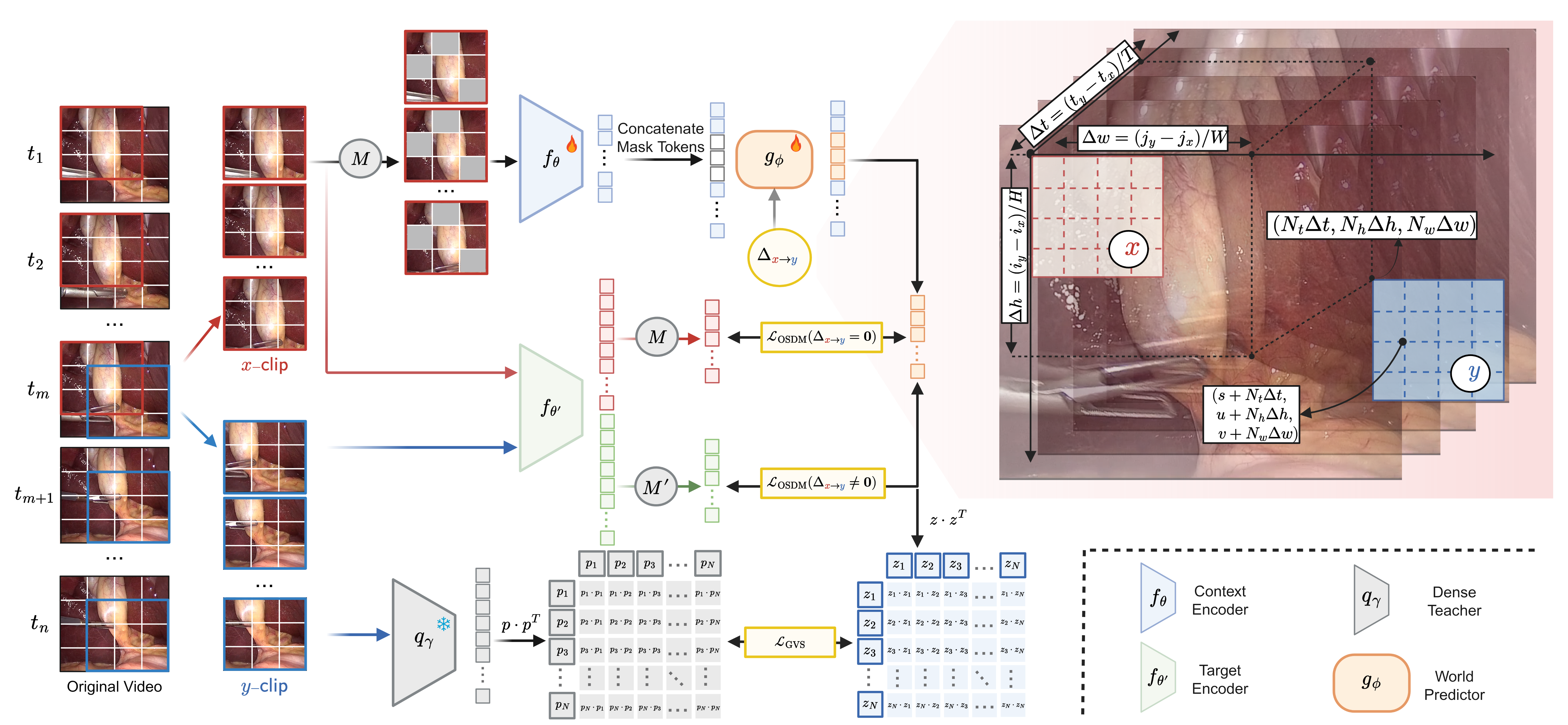}
    \caption{Overview of the SurgWorld framework. Given a context clip $x$ and a target clip $y$ sampled from the same surgical video, SurgWorld encodes the context and predicts the target latent state conditioned on their relative shift $\Delta_{x\rightarrow y}$. Training is driven by two complementary objectives: Out-of-Sight Dynamics Modeling (OSDM) and Geometric Visual Similarity (GVS) loss. $M$ denotes the masking strategy. For clarity, the framework is presented in a simplified form to highlight its key methodological innovations.} 
    \label{fig:mainfig}
\end{figure*}

\section{Method}
This section presents SurgWorld, a world-model-inspired framework for video representation learning.
We propose Out-of-Sight Dynamics Modeling (OSDM), a pretraining task that aims to encode the geometric structure and temporal dynamics of surgical scenes by learning to predict unseen visual content in latent representation space.
The core idea is to train a model to predict how surgical scenes evolve across various spatio-temporal shifts.
After introducing the basic architecture, we first describe the details of the OSDM task.
Then, we propose Geometric Visual Similarity (GVS) loss to enhance the dense perception ability of SurgWorld. 
Finally, we present a unified framework that integrates the complementary strengths of proposed training signals, yielding a powerful agglomerative surgical foundation model that produces semantically rich and transferable representations.

\subsection{Basic Architecture}

Here, we introduce our basic framework, built upon the video-based joint-embedding predictive architecture, which performs masked denoising in representation space.
Specifically, the framework comprises two encoders: a context encoder \(f_{\theta}\) and a target encoder \(f_{\theta'}\), which map the context \(x\) and target \(y\) to their latent representations \(h_x\) and \(h_y\), respectively. 
Conditioned on the multi-block masking configuration, a predictor \(g_{\phi}\) transforms the context representation \(h_x\) into an estimated target representation. 
The prediction objective is then computed entirely in the latent space:
\begin{equation}
\underset{\theta,\phi,m}{\operatorname{minimize}}
\quad
\left\|
g_{\phi}\!\left(m, f_{\theta}(x)\right)
-
\operatorname{sg}\!\left(f_{\theta'}(y)\right)
\right\|_{1}
\end{equation}
where $m$ denotes the learnable mask tokens used to replace the dropped patches.
In practice, following \cite{assran2025v}, the parameters \(\theta'\) of the target encoder are updated as an exponential moving average (EMA) of the context encoder parameters \(\theta\).

\subsection{Out-of-Sight Dynamics Modeling}

Surgeons often infer hidden anatomy and evolving instrument–tissue interactions from limited operative views.
Motivated by this observation, we develop an OSDM task that predicts the latent representation of an unseen target region conditioned on its relative spatio-temporal shifts from an observed context.
As illustrated in Fig.~\ref{fig:mainfig}, we randomly crop two related clips from the same original surgical video $V$, serving as context $x$ and target $y$, respectively. 
SurgWorld learns to predict $y$ from $x$ given their relative information $\Delta_{x \rightarrow y}$. 
The spatio-temporal shift is defined as the proportional relative displacement between the two cropped clips:
\begin{equation}
\resizebox{0.9\linewidth}{!}{$\displaystyle
\Delta_{x \to y}
=
(\Delta t,\Delta h,\Delta w)
=
\left(
\frac{t_y-t_x}{T},
\frac{i_y-i_x}{H},
\frac{j_y-j_x}{W}
\right)
$}
\end{equation}
Here we assume that the context $x$ and target $y$ each contain $T$ frames with spatial resolution $H \times W$, starting at $(t_x,i_x,j_x)$ and $(t_y,i_y,j_y)$ in the original video.
To encode target-token locations relative to the context clip, we establish a coordinate system over the context spatio-temporal token grid, with coordinates ranging from $(0,0,0)$ to $(N_t-1, N_h-1, N_w-1)$. $N_t$, $N_h$ and $N_w$ denote the numbers of tokens along the temporal, height, and width dimensions, respectively.
The coordinate of a target token indexed by temporal position \(s\), row \(u\), and column \(v\) is given by 
\begin{equation}
\resizebox{0.9\linewidth}{!}{$\displaystyle
\phi_{x \to y}(s,u,v)
=
\left(
s + N_t \Delta t,\,
u + N_h \Delta h,\,
v + N_w \Delta w
\right)
$}
\end{equation}
where \(s \in [0, N_t - 1]\), \(u \in [0, N_h - 1]\), and \(v \in [0, N_w - 1]\).
Essentially, $\phi_{x \to y}(s,u,v)$ describes the target's location from the context's perspective. 
To condition the predictor on derived relative spatio-temporal positions, we extend 1D RoPE to 3D by partitioning the feature channels into temporal, height, and width groups and applying axis-specific rotations.
The predictor output is given by:
\begin{equation}
\begin{aligned}
\hat{h}_{y}
&=
g_{\phi}(h_{x}; \Delta_{x \to y}) \\
&=
g_{\phi}^{\mathrm{RoPE}}
\left(
(h_x,p_x)
\oplus
\{m, \phi_{x \to y}(s,u,v)\}_{s,u,v}
\right).
\end{aligned}
\end{equation}
Here, $p_x$ denotes the spatio-temporal coordinates of the visible context tokens. The loss function is given by:
\begin{equation}
\mathcal{L}_{\mathrm{OSDM}}(x,y)
=
\left\|
g_{\phi}(f_{\theta}(x);\Delta_{x \to y})
-
f_{\theta'}(y)
\right\|_1.
\end{equation}

\subsection{Geometric Visual Coherence Alignment}
Beyond inferring out-of-sight surgical states, we introduce a Geometric Visual Similarity (GVS) loss that distills spatially structured relationships from a frozen vision teacher, improving geometric consistency and local coherence for fine-grained perception.
Specifically, we adopt a pre-trained vision foundation model $q_{\gamma}$ \cite{simeoni2025dinov3} as the frozen teacher. Given a frame partitioned into $P$ patches, the predicted masked-patch features are combined with the encoded visible-patch features to form the complete student representation $z_s$. 
The same frame is independently encoded by $q_{\gamma}$ to obtain the corresponding teacher representation $z_t$.
We then compute the pairwise visual similarity matrices as
\begin{equation}
\mathbf{S}_s
=
\bar{\mathbf{Z}}_s\bar{\mathbf{Z}}_s^{\top},
\qquad
\mathbf{S}_t
=
\bar{\mathbf{Z}}_t\bar{\mathbf{Z}}_t^{\top},
\end{equation}
where \(\mathbf{S}_s,\mathbf{S}_t\in\mathbb{R}^{P\times P}\), and each entry represents the cosine similarity between two $L_2$-normalized patch-level features.
We then define the geometric visual similarity loss by aligning the pairwise visual relationships of SurgWorld with those of the frozen teacher:
\begin{equation}
\mathcal{L}_{\mathrm{GVS}}
=
\frac{1}{N_t}
\sum_{\tau=0}^{N_t-1}
\left\|
\mathbf{S}_{s}^{(\tau)}
-
\mathbf{S}_{t}^{(\tau)}
\right\|_{F}^{2},
\end{equation}
where $N_t$ is the number of tublets.
The loss transfers the teacher’s spatial structure to SurgWorld, improving geometric consistency and local visual coherence.

\begin{table*}[!t]
    \centering
    \footnotesize
    \setlength{\tabcolsep}{3pt}
    
    \caption{Overview and statistics of the surgical datasets used for SurgWorld pre-training and downstream evaluation. Video datasets are reported as the number of clips, with frame counts in parentheses, whereas image datasets are reported by the number of images. Datasets marked with * contribute their training splits to SurgWorld pre-training.}
    \begin{tabular*}{\textwidth}{
        @{\extracolsep{\fill}}
        l c c c c c
        @{}
    }
        \toprule

        \multirow{2}{*}{\textbf{Dataset}}
        & \multirow{2}{*}{\textbf{Surgical Procedure}}
        & \multicolumn{3}{c}{\textbf{Data Statistics}}
        & \multirow{2}{*}{\textbf{Task}} \\

        \cmidrule(lr){3-5}

        &
        & \textbf{Train}
        & \textbf{Val+Test}
        & \textbf{All}
        & \\

        \midrule



        Cholec80$^{*}$~\cite{twinanda2016endonet}
        & Laparoscopic Cholecystectomy
        & 40 (86,344)
        & 35 (88,459)
        & 75 (174,803)
        & Surgical Workflow Recognition \\

        \cmidrule(lr){1-6}

        M2CAI16-Workflow$^{*}$~\cite{stauder2016tum}
        & Laparoscopic Cholecystectomy
        & 27 (67,578)
        & 14 (26,957)
        & 41 (94,535)
        & Surgical Workflow Recognition \\

        \cmidrule(lr){1-6}

        AutoLaparo$^{*}$~\cite{wang2022autolaparo}
        & Laparoscopic Hysterectomy
        & 10 (40,211)
        & 11 (39,479)
        & 21 (79,690)
        & Surgical Workflow Recognition \\

        \cmidrule(lr){1-6}

        PMLR50~\cite{guo2025surgical}
        & Laparoscopic Liver Resection
        & 35 (17,378)
        & 15 (7,659)
        & 50 (25,037)
        & Surgical Workflow Recognition \\

        \cmidrule(lr){1-6}

        CATARACTS~\cite{al2019cataracts}
        & Cataract Surgery
        & 25 (16,953)
        & 25 (15,861)
        & 50 (32,814)
        & Surgical Workflow Recognition \\

        \cmidrule(lr){1-6}

        Cataract101$^{*}$~\cite{schoeffmann2018cataract}
        & Cataract Surgery
        & 50 (26,606)
        & 50 (23,270)
        & 100 (49,876)
        & Surgical Workflow Recognition \\

        \cmidrule(lr){1-6}

        Cataract-21~\cite{primus2018frame}
        & Cataract Surgery
        & 18 (7,716)
        & 3 (937)
        & 21 (8,653)
        & Surgical Workflow Recognition \\

        \cmidrule(lr){1-6}
        


        SurgicalActions160~\cite{schoeffmann2018video}
        & Gynecologic Laparoscopy
        & 112 (13,297)
        & 48 (5,724)
        & 160 (19,021)
        & Surgical Action Recognition \\

        \cmidrule(lr){1-6}

        Cholec80-CVS~\cite{rios2023cholec80}
        & Laparoscopic Cholecystectomy
        & 352 (55,851)
        & 208 (22,620)
        & 560 (78,471)
        & Surgical Skill Assessment \\

        \cmidrule(lr){1-6}

        Endoscapes-CVS~\cite{murali2023latent}
        & Laparoscopic Cholecystectomy
        & 120 (36,694)
        & 81 (22,119)
        & 201 (58,813)
        & Surgical Skill Assessment \\

        \cmidrule(lr){1-6}

        CholecT50~\cite{nwoye2022rendezvous}
        & Laparoscopic Cholecystectomy
        & 45 (90,489)
        & 5 (10,374)
        & 50 (100,863)
        & Surgical Triplet Recognition \\

        \cmidrule(lr){1-6}



        Kvasir-SEG~\cite{jha2019kvasir}
        & Colonoscopy
        & 900
        & 100
        & 1,000
        & Polyp Segmentation \\

        \cmidrule(lr){1-6}

        CVC-ClinicDB~\cite{bernal2015wm}
        & Colonoscopy
        & 550
        & 62
        & 612
        & Polyp Segmentation \\

        \cmidrule(lr){1-6}

        CVC-300~\cite{vazquez2017benchmark}
        & Colonoscopy
        & --
        & 60
        & 60
        & Polyp Segmentation \\

        \cmidrule(lr){1-6}

        CVC-ColonDB~\cite{bernal2012towards}
        & Colonoscopy
        & --
        & 380
        & 380
        & Polyp Segmentation \\

        \cmidrule(lr){1-6}

        ETIS-LaribPolypDB~\cite{silva2014toward}
        & Colonoscopy
        & --
        & 196
        & 196
        & Polyp Segmentation \\

        \cmidrule(lr){1-6}
        CholecSeg8k~\cite{hong2020cholecseg8k}
        & Laparoscopic Cholecystectomy
        & 6,080
        & 2,000
        & 8,080
        & Surgical Semantic Segmentation \\

        \cmidrule(lr){1-6}
        HeiChole$^{*}$~\cite{wagner2023comparative}
        & Laparoscopic Cholecystectomy
        & 24 (55,139)
        & --
        & --
        & -- \\

        \cmidrule(lr){1-6}
        PitVis$^{*}$~\cite{das2025pitvis}
        & Endoscopic Pituitary Surgery 
        & 25 (120,018)
        & --
        & --
        & -- \\

        \cmidrule(lr){1-6}
        PSI-AVA$^{*}$~\cite{valderrama2022towards}
        & Robot-Assisted Radical Prostatectomy 
        & 8 (73,618)
        & --
        & --
        & -- \\

        \cmidrule(lr){1-6}
        BernBypass70$^{*}$~\cite{lavanchy2024challenges}
        & Laparoscopic Roux-en-Y Gastric Bypass 
        & 70 (316,676)
        & --
        & --
        & -- \\

        \cmidrule(lr){1-6}
        StrasBypass70$^{*}$~\cite{lavanchy2024challenges}
        & Laparoscopic Roux-en-Y Gastric Bypass 
        & 70 (464,989)
        & --
        & --
        & -- \\

        \bottomrule
    \end{tabular*}
    \label{tab:dataset}
    
\end{table*}

\subsection{Unified Agglomerative Training Framework}
Building on OSDM and GVS, we unify their complementary supervision through shared contexts, targets, and actions, enabling both objectives to be optimized within a single forward pass in an agglomerative training framework.
As illustrated in Fig.~\ref{fig:mainfig}, 
we sample two distinct, equally sized target clips $y_1$ and $y_2$ from an input video $V$, and mask each target to construct the context clips $x_1$ and $x_2$, and then encoded as
\begin{equation}
h_{x_i}
=
f_{\theta}(x_i),
h_{y_i}
=
f_{\theta'}(y_i),
z_{i}
=
q_{\gamma}(y_i),
i=1,2.
\end{equation}
Here, $h_{y_i}$ denotes the target representation produced by the EMA teacher encoder, while $z_i$ denotes the representation extracted by the frozen pre-trained vision teacher.

Since \(x_1\) is obtained by masking \(y_1\), predicting \(y_1\) from \(x_1\), conditioned on $M_1$, constitutes a special case of out-of-sight dynamics modeling in which the relative spatio-temporal shift satisfies $\Delta_{x_1 \rightarrow y_1}=\mathbf{0}$. 
To further enhance the geometric coherence of the learned representations, we construct the geometric visual similarity matrices $S_{s_{1}}$ and $S_{t_{1}}$ from the normalized student representation and the representation extracted by the frozen visual teacher, respectively:
{
\begin{equation}
\begin{aligned}
\mathcal{L}_{1\rightarrow1}
&=
\mathcal{L}_{\mathrm{OSDM}}^{1\rightarrow1}
+
\lambda_{\mathrm{GVS}}\mathcal{L}_{\mathrm{GVS}}^{1}
\\
&=
\sum_{p\in\mathcal{M}_1}
\left\|
g_{\phi}
\left(
h_{x_1};\mathcal{M}_1
\right)_p
-
h_{y_1,p}
\right\|_1
\\
&\quad+
\frac{\lambda_{\mathrm{GVS}}}{N_t}
\sum_{\tau=0}^{N_t-1}
\left\|
\mathbf{S}_{s_{1}}^{(\tau)}
-
\mathbf{S}_{t_{1}}^{(\tau)}
\right\|_F^2 .
\end{aligned}
\end{equation}
}
Moreover, $x_1$ and $y_2$ are related by the nonzero spatio-temporal shift $\Delta_{x_1\rightarrow y_2}$. Accordingly, the predictor can simultaneously predict $y_2$ from $x_1$, conditioned on this relative shift, thereby realizing out-of-sight dynamics modeling within the same forward pass:
\begin{equation}
\begin{aligned}
\mathcal{L}_{1 \rightarrow 2}
&=
\sum_{p \in M_2}
\left\|
g_{\phi}(h_{x_1};M_2, \Delta_{x_1 \to y_2})_p
-
h_{y_2,p}
\right\|_1 \\
&\quad+
\frac{\lambda_{\mathrm{GVS}}}{N_t}
\sum_{\tau=0}^{N_t-1}
\left\|
\mathbf{S}_{s_{1\rightarrow 2}}^{(\tau)}
-
\mathbf{S}_{t_{2}}^{(\tau)}
\right\|_F^2 .
\end{aligned}
\end{equation}

Similarly, \(\mathcal{L}_{2 \rightarrow 2}\) and \(\mathcal{L}_{2 \rightarrow 1}\) can also be computed between \((x_2,y_2)\) and \((x_2,y_1)\), respectively.
Note that the context features $h_{x_1}$, $h_{x_{2}}$ and target features $h_{y_1}$, $h_{y_2}$ are reused in the above loss computation. 
Finally, the model is trained with the unified objective:
\begin{equation}
\mathcal{L}_{\mathrm{total}}
=
\mathcal{L}_{1 \rightarrow 1}
+
\mathcal{L}_{2 \rightarrow 2}
+
\mathcal{L}_{1 \rightarrow 2}
+
\mathcal{L}_{2 \rightarrow 1}
.  
\end{equation}
Here, $\lambda_{\mathrm{GVS}}$ is set to $2.0$. 
Based on this unified objective, SurgWorld learns a predictive and geometry-aware representation of the surgical environment, enabling it to infer out-of-sight surgical states while capturing spatially structured scene evolution and preserving coherent dense visual representations.

\section{Experiments and Results}

\subsection{Datasets} 
To support comprehensive pretraining and evaluation, we systematically curated a diverse collection of publicly available surgical video and image datasets. Specifically, SurgWorld was pretrained on nine publicly available surgical video datasets. We further evaluated SurgWorld on 17 downstream benchmarks spanning surgical workflow recognition, fine-grained action understanding, and dense visual perception. To prevent data leakage, we ensured that no video used during pretraining appeared in the test split of any downstream benchmark. For datasets used in both pretraining and downstream evaluation, all test-set videos were explicitly excluded from the pretraining corpus. Table~\ref{tab:dataset} summarizes the pretraining and evaluation datasets, including the surgical procedures covered, split-specific sample statistics, and corresponding task categories.

\subsection{Comparison Methods}
We compare SurgWorld against 12 representative pretrained models from three categories: general-domain self-supervised models, surgical-domain visual foundation models, and surgical vision-language foundation models. These baselines cover diverse pretraining paradigms, including masked reconstruction, contrastive learning, predictive representation learning and multimodal alignment.

\subsubsection{General-domain self-supervised models}
\begin{itemize}
\item \textbf{DINOv3}\cite{simeoni2025dinov3} learns transferable image representations through large-scale teacher--student self-distillation; we evaluate its ViT-L and ViT-H variants.
\item \textbf{VideoMAEv2}\cite{wang2023videomae} employs dual masking for scalable masked video reconstruction; we evaluate its ViT-L and ViT-G variants.
\end{itemize}

\begin{table*}[t]
    \centering
    \footnotesize
    \setlength{\tabcolsep}{3pt}
    
    \caption{Performance on seven downstream surgical workflow recognition benchmarks using attentive probing. Results are reported in accuracy and F1 score. The best two results are bold-faced and underlined, respectively. $^{*}$ denotes significant improvements over the strongest baseline on both metrics of the marked dataset under two-sided Wilcoxon signed-rank tests ($p<0.05$). $^{\dagger}$ denotes the inclusion of private data in pre-training.} 
    \begin{tabular}{l l l | *{14}{c}}
        \toprule
        \multirow{2}{*}{\textbf{Models}} 
        & \multirow{2}{*}{\textbf{Pretrain. data}} 
        & \multirow{2}{*}{\textbf{Backbone}} 
        & \multicolumn{2}{c}{\textbf{Cholec80$^{*}$}} 
        & \multicolumn{2}{c}{\textbf{AutoLaparo$^{*}$}} 
        & \multicolumn{2}{c}{\textbf{M2CAI16$^{*}$}} 
        & \multicolumn{2}{c}{\textbf{PMLR50$^{*}$}} 
        & \multicolumn{2}{c}{\textbf{Cataract$^{*}$}} 
        & \multicolumn{2}{c}{\textbf{Cataract21$^{*}$}} 
        & \multicolumn{2}{c}{\textbf{Cataract101$^{*}$}} \\
        \cmidrule(lr){4-5}
        \cmidrule(lr){6-7}
        \cmidrule(lr){8-9}
        \cmidrule(lr){10-11}
        \cmidrule(lr){12-13}
        \cmidrule(lr){14-15}
        \cmidrule(lr){16-17}
        & & 
        & Acc & F1
        & Acc & F1
        & Acc & F1
        & Acc & F1
        & Acc & F1
        & Acc & F1
        & Acc & F1 \\
        \midrule

        \multirow{2}{*}{DINOv3} 
        & \multirow{2}{*}{LVD-1689M} 
        & ViT-L 
        & \underline{85.16} & \underline{77.56} 
        & 76.28 & 67.43 
        & \underline{83.19} & \underline{70.18} 
        & 67.67 & 53.72 
        & 75.84 & \underline{70.18} 
        & 79.91 & \underline{75.11} 
        & \underline{87.59} & \underline{83.43} \\

        & & ViT-H 
        & 84.95 & 76.75 
        & 76.85 & \underline{67.55} 
        & 82.52 & 69.86 
        & 72.32 & 54.99 
        & 72.04 & 67.47 
        & 74.81 & 68.15 
        & 85.29 & 80.43 \\

        \multirow{2}{*}{VideoMAEv2} 
        & \multirow{2}{*}{7500h} 
        & ViT-L 
        & 80.30 & 72.84 
        & 69.88 & 58.73 
        & 75.27 & 62.24 
        & \underline{75.29} & \underline{56.62} 
        & 71.26 & 63.65 
        & \underline{83.44} & 73.73 
        & 85.44 & 79.44 \\

        & & ViT-G
        & 76.69 & 68.20 
        & 61.38 & 52.18 
        & 71.18 & 59.11 
        & 72.88 & 52.24 
        & 73.85 & 64.81 
        & 60.46 & 53.38 
        & 78.94 & 71.13 \\

        \midrule

        SurgMotion 
        & 3658h$^{\dagger}$ (1fps) 
        & ViT-L 
        & 80.55 & 73.40 
        & 70.44 & 62.06 
        & 77.88 & 68.21 
        & 71.88 & 55.14 
        & \underline{76.14} & 68.51 
        & 72.21 & 57.64 
        & 81.81 & 73.44 \\

        EndoViT 
        & 194h (1fps) 
        & ViT-B
        & 81.35 & 73.47 
        & 69.83 & 60.50 
        & 76.33 & 62.77 
        & 65.35 & 46.76 
        & 64.96 & 55.74 
        & 66.06 & 57.76 
        & 82.83 & 77.71 \\

        GSViT 
        & 680h (30fps) 
        & EffViT-M5
        & 48.03 & 23.67 
        & 36.24 & 26.72 
        & 34.32 & 19.00 
        & 65.29 & 20.84 
        & 21.42 & 7.10 
        & 34.66 & 4.63 
        & 22.71 & 4.81 \\ 

        Endo-Mamba 
        & 103h$^{\dagger}$ (30fps)
        & Mamba-S 
        & 52.23 & 39.96 
        & 43.50 & 38.49 
        & 44.29 & 32.92 
        & 63.95 & 30.74 
        & 20.98 & 12.18 
        & 31.02 & 7.30 
        & 32.14 & 11.43 \\

        EndoSSL 
        & 3028h$^{\dagger}$ (1fps) 
        & ViT-L
        & 47.59 & 36.13 
        & 42.99 & 36.25 
        & 43.05 & 31.05 
        & 62.70 & 33.94 
        & 22.79 & 12.63 
        & 32.42 & 5.24 
        & 31.49 & 9.32 \\

        EndoFM 
        & 46h$^{\dagger}$ (30fps) 
        & ViT-B
        & 83.40 & 76.50 
        & 73.47 & 62.12 
        & 78.64 & 66.69 
        & 66.52 & 48.32 
        & 66.22 & 60.27 
        & 76.51 & 70.50 
        & 82.24 & 77.08 \\

        SelfSupSurg 
        & 56h (5fps)
        & ResNet50 
        & 46.99 & 25.17 
        & 40.11 & 25.99 
        & 38.95 & 21.89 
        & 47.35 & 18.00 
        & 26.87 & 8.77 
        & 39.67 & 7.05 
        & 29.15 & 4.55 \\

        SurgeNet 
        & 1308h$^{\dagger}$ (1fps) 
        & ConvNeXt 
        & 80.59 & 73.90 
        & \underline{77.32} & 64.85 
        & 79.52 & 66.61 
        & 68.77 & 49.81 
        & 59.15 & 48.55 
        & 50.96 & 37.22 
        & 67.93 & 56.32 \\

        \midrule

        SurgVLP 
        & 300h$^{\dagger}$ (2fps) 
        & ResNet50 
        & 80.28 & 72.21 
        & 71.49 & 61.11 
        & 76.04 & 61.78 
        & 65.45 & 47.30 
        & 63.41 & 56.38 
        & 72.38 & 56.55 
        & 77.17 & 71.16 \\

        PeskaVLP 
        & 300h$^{\dagger}$ (2fps) 
        & ResNet50 
        & 82.03 & 74.75 
        & 76.11 & 66.70 
        & 78.07 & 64.75 
        & 66.27 & 47.90 
        & 65.45 & 59.13 
        & 77.79 & 62.83 
        & 82.21 & 77.33 \\

        \midrule
        
        SurgWorld 
        & 348h (1fps)
        & ViT-L 
        & \textbf{89.42} & \textbf{81.64} 
        & \textbf{83.04} & \textbf{72.62} 
        & \textbf{84.49} & \textbf{77.06} 
        & \textbf{82.31} & \textbf{67.68} 
        & \textbf{83.02} & \textbf{78.20} 
        & \textbf{91.94} & \textbf{90.13} 
        & \textbf{92.68} & \textbf{90.98} \\

        \bottomrule
    \end{tabular}
    \label{tab:workflow}
\end{table*}

\begin{table*}[t]
    \centering
    \footnotesize
    \setlength{\tabcolsep}{3pt}
    
    \caption{Performance on four surgical fine-grained action understanding benchmarks using attentive probing. mAP is reported for CholecT50 and Acc / F1 is reported for the rest of the datasets. The best two results are bold-faced and underlined, respectively.}
    \scalebox{1.0}{
    \begin{tabular}{l l | c c c c | c c | c c c | c c c}
        \toprule
        \multirow{2}{*}{\textbf{Models}} 
        & \multirow{2}{*}{\textbf{Backbone}} 
        & \multicolumn{2}{c}{\textbf{Cholec80CVS$^{*}$}}
        & \multicolumn{2}{c|}{\textbf{EndoscapesCVS}} 
        & \multicolumn{2}{c|}{\textbf{SurgicalActions160}} 
        & \multicolumn{6}{c}{\textbf{CholecT50$^{*}$}} \\
        \cmidrule(lr){3-4}
        \cmidrule(lr){5-6}
        \cmidrule(lr){7-8}
        \cmidrule(lr){9-14}
        &
        & \textbf{Acc} & \textbf{F1}
        & \textbf{Acc} & \textbf{F1}
        & \textbf{Acc} & \textbf{F1}
        & \textbf{AP$_i$} & \textbf{AP$_v$} 
        & \textbf{AP$_t$} & \textbf{AP$_{iv}$} 
        & \textbf{AP$_{it}$} & \textbf{AP$_{ivt}$}
        \\
        \midrule

        \multirow{2}{*}{DINOv3}
        & ViT-L
        & 51.67 & 27.09
        & \underline{58.20} & \underline{38.47}
        & 35.42 & 32.58
        & \underline{86.51} & 56.25 & \underline{45.69} & 36.94 & 38.62 & 33.85 \\

        & ViT-H
        & 46.67 & 16.33
        & 57.36 & 34.62 
        & 20.83 & 13.19
        & 86.26 & \underline{56.27} & 45.09 & \underline{37.44} & \underline{39.31} & \underline{35.07} \\

        \multirow{2}{*}{VideoMAEv2}
        & ViT-L
        & 45.00 & 29.87
        & 52.81 & 29.55
        & \underline{58.33} & \underline{57.44}
        & 80.90 & 50.39 & 42.77 & 32.11 & 32.19 & 28.46 \\

        & ViT-G
        & 49.17 & \underline{35.38}
        & 54.69 & 25.85
        & 27.08 & 23.01
        & 69.68 & 44.94 & 41.73 & 27.38 & 28.86 & 27.07 \\

        \midrule

        SurgMotion
        & ViT-L
        & 45.00 & 23.32
        & 54.64 & 34.18
        & 47.92 & 47.18
        & 71.58 & 45.48 & 43.33 & 30.10 & 36.05 & 33.08 \\

        EndoViT
        & ViT-B
        & 46.67 & 18.36
        & 57.80	& 33.69
        & 22.92 & 22.57
        & 70.71 & 46.59 & 42.57 & 29.78 & 31.00 & 29.62 \\

        GSViT
        & EffViT-M5
        & 39.17 & 13.86
        & 57.31 & 18.22
        & 18.75 & 10.18
        & 35.09 & 22.98 & 16.60 & 12.97 & 9.68 & 11.65 \\

        Endo-Mamba
        & Mamba-S
        & 37.50 & 14.76
        & 57.31 & 18.22
        & 14.58 & 10.61
        & 41.75 & 27.29 & 22.27 & 16.15 & 18.06 & 16.64 \\

        EndoSSL
        & ViT-L
        & 42.37 & 16.35
        & 57.14 & 20.12
        & 20.83 & 16.68
        & 39.78 & 27.56 & 21.65 & 17.23 & 14.53 & 14.28 \\

        EndoFM
        & ViT-B
        & 50.83 & 28.15
        & 53.47 & 32.27
        & 43.75 & 42.02
        & 77.97 & 48.60 & 42.58 & 30.72 & 32.79 & 29.02 \\

        SelfSupSurg
        & ResNet50
        & 43.33 & 15.77
        & 44.19 & 23.87
        & 12.50 & 9.14
        & 34.64 & 21.43 & 18.45 & 12.78 & 10.28 & 9.08 \\

        SurgeNet
        & ConvNeXt
        & \underline{52.83} & 29.88
        & 53.19 & 33.01
        & 43.75 & 43.14
        & 68.31 & 44.53 & 39.09 & 30.10 & 27.05 & 23.62 \\

        \midrule

        SurgVLP
        & ResNet50
        & 38.33 & 22.05
        & 47.42 & 30.95
        & 43.75 & 45.18
        & 68.39 & 43.18 & 39.62 & 26.98 & 28.65 & 25.00 \\

        PeskaVLP
        & ResNet50
        & 50.00 & 29.78
        & 40.03 & 30.15
        & 52.08 & 48.21
        & 76.06 & 50.18 & 39.74 & 31.88 & 32.73 & 29.30 \\

        \midrule

        \textbf{SurgWorld}
        & \textbf{ViT-L}
        & \textbf{53.39} & \textbf{40.72}
        & \textbf{59.36} & \textbf{41.71}
        & \textbf{60.42} & \textbf{59.10}
        & \textbf{92.04} & \textbf{60.25} & \textbf{52.28} 
        & \textbf{42.76} & \textbf{44.69} & \textbf{44.32} \\

        \bottomrule
    \end{tabular}
    }
    \label{tab:fine-grained}
\end{table*}

\subsubsection{Surgical-domain visual foundation models}
\begin{itemize}
\item \textbf{SurgMotion}\cite{wu2026unisurg} extends the V-JEPA paradigm with motion-guided masked prediction, spatiotemporal affinity distillation, and feature-diversity regularization.
\item \textbf{EndoViT}\cite{batic2024endovit} uses masked autoencoding on large-scale endoscopic images to learn spatial representations without explicitly modeling temporal dynamics.
\item \textbf{GSViT}\cite{schmidgall2024general} adopts an EfficientViT architecture and forward frame prediction to learn surgical representations from future-frame supervision.
\item \textbf{EndoMamba}\cite{tian2025endomamba} combines bidirectional spatial Mamba blocks and causal temporal Mamba blocks within a hierarchical self-supervised pretraining framework.
\item \textbf{EndoSSL}\cite{hirsch2023self} employs Masked Siamese Networks to learn image-level representations from large-scale endoscopic video data.
\item \textbf{EndoFM}\cite{wang2023foundation} learns endoscopic video representations through self-supervised spatio-temporal matching.
\item \textbf{SelfSupSurg}\cite{ramesh2023dissecting} investigates the transferability of established image-level self-supervised learning methods to surgical vision; we adopt its DINO-pretrained ResNet-50 checkpoint.
\item \textbf{SurgeNet}\cite{de2025scaling} scales DINO-based self-supervised pretraining across diverse surgical procedures using a ConvNeXtV2 encoder.
\end{itemize}

\subsubsection{Surgical vision-language foundation models}
\begin{itemize}
\item \textbf{SurgVLP}\cite{yuan2025learning} learns multimodal surgical representations by contrastively aligning video clips with narration-derived text.
\item \textbf{PeskaVLP}\cite{yuan2024procedure} combines video-text alignment, visual self-supervision, and hierarchical procedure-aware textual supervision to capture surgical semantics across multiple temporal scales.
\end{itemize}

\subsection{Implementation Details}

For pretraining, we initialized SurgWorld entirely from scratch, employing random initialization for all components.
SurgWorld uses ViT-Large~\cite{dosovitskiy2020image} for both the context and target encoders. The narrow ViT predictor comprises 12 Transformer blocks with an embedding dimension of 384. Both encoders tokenize the input using tubelets of size $2\times16\times16$. All videos are sampled at 1 frame per second (fps) with an input resolution of $256\times256$ pixels. We follow the multi-block masking strategy in~\cite{assran2025v} to promote spatio-temporal consistency during pre-training. The target encoder parameters are updated using an exponential moving average (EMA) of the context encoder parameters, with a fixed momentum of $\alpha=0.99925$.

Pretraining proceeds in two stages, each with 300 iterations per epoch. Stage 1 uses 16-frame clips for 250 epochs (75,000 iterations). The learning rate increases linearly from $1\times10^{-4}$ to $5.25\times10^{-4}$ over the first 40 epochs (12,000 iterations) and remains constant thereafter. The GVS objective is activated after two epochs (600 iterations), followed by a linear warmup over the next eight epochs (2,400 iterations). Stage 2 uses 64-frame clips for a 40-epoch cooldown (12,000 iterations). During this stage, the learning rate decreases linearly from $5.25\times10^{-4}$ to $1\times10^{-6}$. The temporal displacement $\Delta t$ is sampled randomly from $\{0,\ldots,8\}$ frames in Stage 1 and $\{0,\ldots,16\}$ frames in Stage 2. We optimize the model using AdamW~\cite{loshchilov2017decoupled}, with per-GPU batch sizes of 32 and 16 for Stages 1 and 2, respectively. 

For downstream evaluation, we freeze the pretrained backbone and train task-specific probes. We use a four-block attentive probe with 16 attention heads for surgical workflow recognition and fine-grained action understanding. For dense visual perception, we apply a lightweight linear segmentation probe to the final-layer patch features. All compared methods use the same hyperparameter search space. We report accuracy and macro-F1 for workflow and action recognition. Surgical triplet recognition and segmentation are evaluated using average precision and Dice score, respectively.
All experiments are conducted on four NVIDIA H200 GPUs.

\begin{table}[t]
    \centering
    \footnotesize
    \setlength{\tabcolsep}{1pt}

    \caption{Performance on six surgical segmentation benchmarks (ID and OOD settings for polyp segmentation). Results are reported in Dice score.}
    \scalebox{1.0}{
    \begin{tabular}{l | c c | c c c | c}
        \toprule
        \multirow{2}{*}{\textbf{Models}}
        & \multicolumn{2}{c|}{\textbf{In-Distribution}}
        & \multicolumn{3}{c|}{\textbf{Out-of-Distribution}}
        &  \\
        
        \cmidrule(lr){2-3}
        \cmidrule(lr){4-6}
        & ClinicDB & Kvasir
        & CVC-300 & ColonDB & ETIS
        & CholecSeg8k \\

        \midrule

        DINOv3
        & 78.95 & 82.04
        & 79.47 & 60.11 & 65.22
        & 75.29 \\

        VideoMAE
        & 71.18 & 74.94
        & 71.06 & 50.83 & 51.80
        & 69.49 \\
        
        \midrule

        SurgMotion
        & 63.40 & 67.15
        & 53.52 & 41.25 & 41.37
        & 68.66 \\

        EndoViT
        & 67.09 & 71.36
        & \underline{69.07} & 54.58 & \underline{58.68}
        & \underline{72.59} \\

        GSViT
        & 34.85 & 46.80
        & 20.70 & 24.74 & 19.51
        & 42.90 \\

        Endo-Mamba
        & 32.14 & 40.91
        & 14.93 & 22.93 & 12.52
        & 43.46 \\

        EndoSSL
        & 49.84 & 47.42
        & 26.48 & 33.79 & 28.89
        & 65.62 \\

        EndoFM
        & \underline{72.79} & \underline{75.91}
        & 66.18 & \underline{55.13} & 56.81
        & 67.15 \\

        SelfSupSurg
        & 40.95 & 50.64
        & 22.60 & 27.14 & 21.90
        & 49.39 \\

        SurgeNet
        & 69.93 & 73.20
        & 67.62 & 51.04 & 56.09
        & 69.24 \\

        \midrule
        
        SurgVLP
        & 61.39 & 65.58
        & 57.94 & 40.26 & 40.02
        & 60.98 \\

        PeskaVLP
        & 65.16 & 69.37
        & 57.69 & 44.16 & 49.27
        & 58.90 \\

        \midrule

        \textbf{SurgWorld$^{*}$}
        & \textbf{79.79} & \textbf{83.75}
        & \textbf{79.58} & \textbf{60.93} & \textbf{66.22}
        & \textbf{80.36} \\

        \bottomrule
    \end{tabular}
    }
    \label{tab:segmentation}
\end{table}

\begin{figure}[t]
    \centering
    \includegraphics[width=\columnwidth]{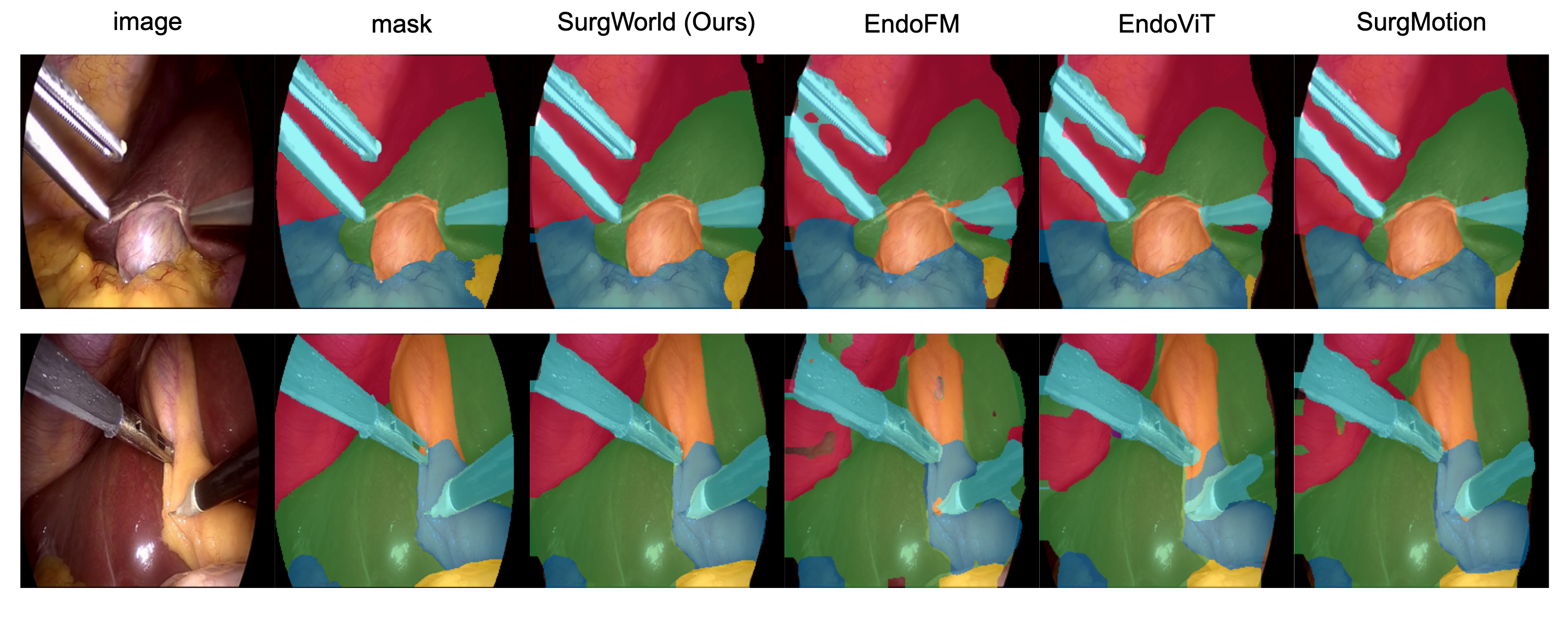}
    \caption{Qualitative comparison of semantic segmentation results on CholecSeg8k.}
    \label{fig:id_ood_seg}
\end{figure}

\begin{figure}[t]
    \centering
    \includegraphics[width=\columnwidth]{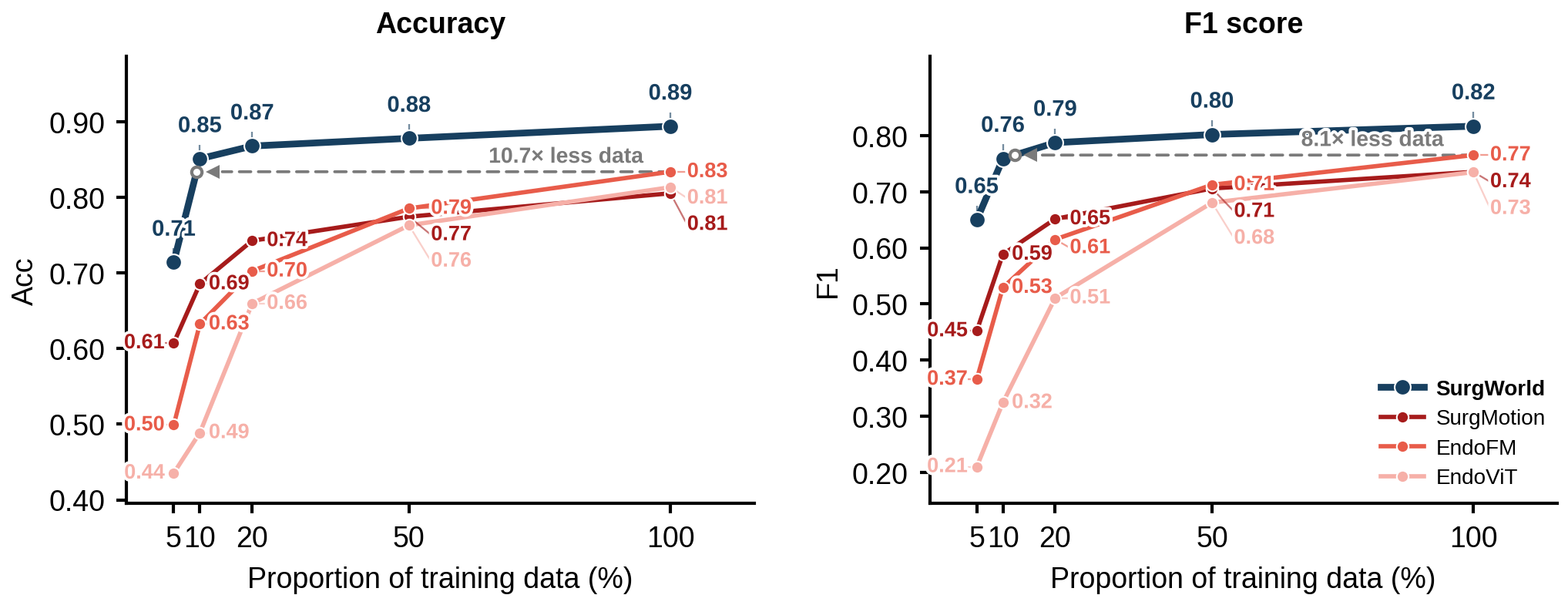}
    \caption{Fine-tuning with 5\%, 10\%, 20\%, 50\% and 100\% training data on Cholec80.}
    \label{fig:result}
\end{figure}

\begin{figure}[t]
    \centering
    \includegraphics[width=\columnwidth]{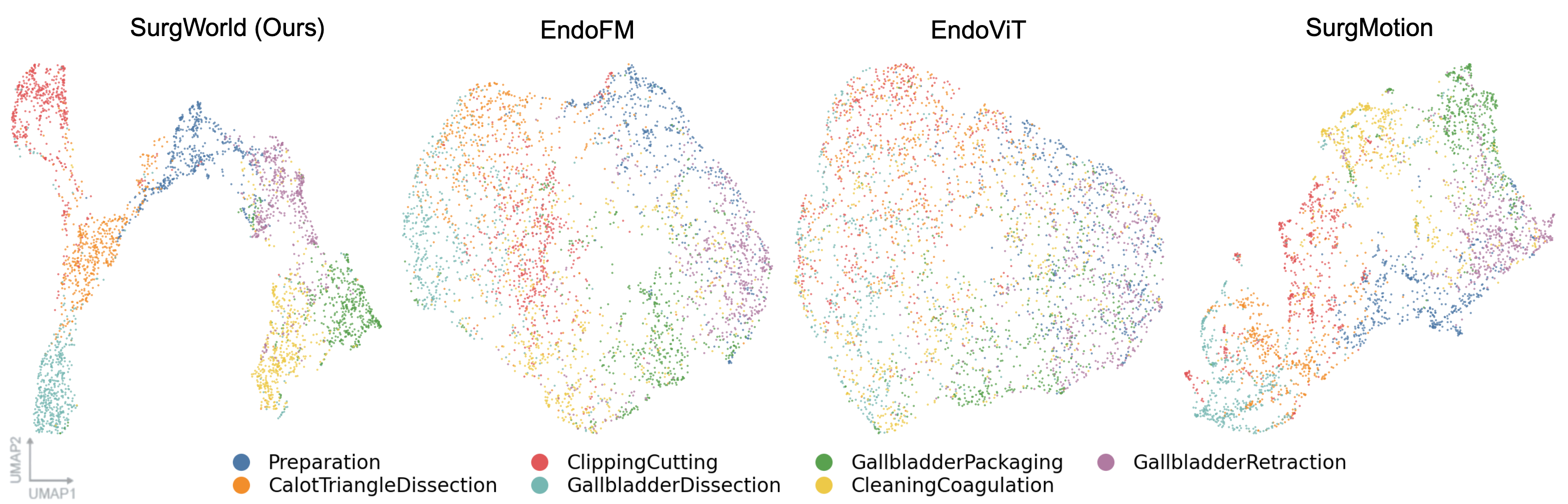}
    \caption{UMAP visualizations of learned representations. Fine-tuning with only 10\% of the labeled data yields seven clearly separated phase clusters.}
    \label{fig:umap}
\end{figure}

\begin{figure}[t]
    \centering
    \includegraphics[width=\columnwidth]{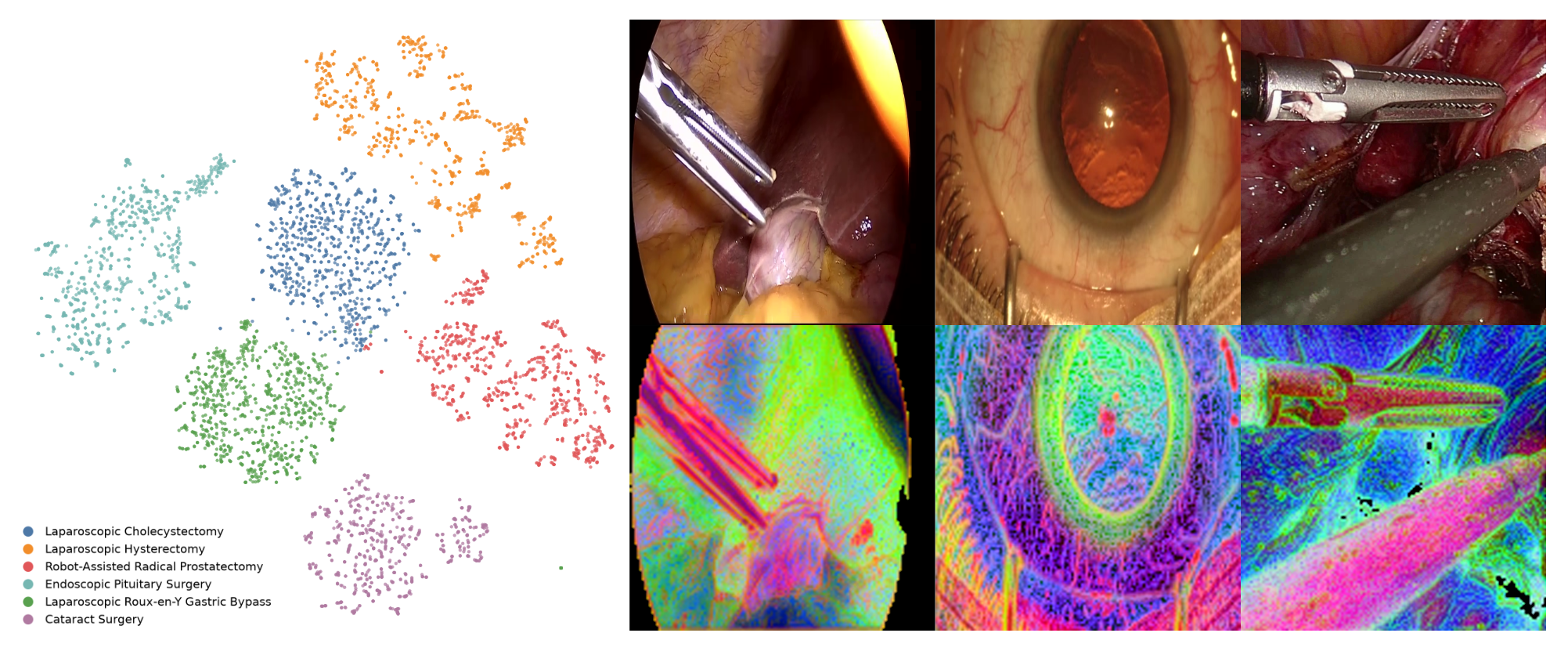}
    \caption{t-SNE visualization of frozen video representations colored by surgical procedure categories (left), and PCA projections of dense feature maps from representative surgical scenes (right).}
    \label{fig:pca-tsne}
\end{figure}

\subsection{SurgWorld produces transferable representations}

\subsubsection{Surgical workflow recognition}
This task aims to identify the current surgical phase from video streams, providing a direct setting for evaluating the temporal and procedural understanding of surgical representations. 
We benchmark SurgWorld on seven workflow recognition datasets (Cholec80 \cite{twinanda2016endonet}, AutoLaparo \cite{wang2022autolaparo}, M2CAI-2016 \cite{stauder2016tum}, PMLR50 \cite{guo2025surgical}, Cataract \cite{al2019cataracts}, Cataract21 \cite{primus2018frame}, Cataract101 \cite{schoeffmann2018cataract}). These datasets span diverse surgical procedures, including cholecystectomy (Cholec80 and M2CAI-2016), hysterectomy (AutoLaparo), laparoscopic liver resection (PMLR50), and cataract surgery (Cataract, Cataract21, and Cataract101).
As shown in Table~\ref{tab:workflow}, SurgWorld consistently outperforms all compared baselines. For example, on Cholec80, SurgWorld achieves an accuracy of 89.42\% and an F1 score of 81.64\%, outperforming the substantially larger DINOv3-H by 4.26 percentage points in accuracy. This substantial improvement suggests that SurgWorld learns representations that are more sensitive to procedure-level temporal context and phase transitions than frame-wise discriminative models, while also highlighting the importance of in-domain surgical pretraining. 
Notably, SurgWorld, pretrained exclusively on 348 hours of publicly available surgical videos, outperforms SurgMotion despite using less than one-tenth of its pretraining data.

\subsubsection{Surgical fine-grained action understanding}
Beyond phase-level workflow recognition, surgical video understanding requires fine-grained reasoning about surgical actions, instruments, anatomical targets, and their interactions. 
Therefore, We evaluate SurgWorld on three complementary tasks: surgical skill assessment, action recognition, and action triplet recognition.
Accurate assessment of Strasberg's Critical View of Safety (CVS) is crucial for reducing the risk of injury to critical anatomical structures and promoting procedural standardization, thereby supporting intraoperative monitoring and postoperative quality evaluation. We evaluate this ability on Cholec80-CVS\cite{rios2023cholec80} and Endoscapes-CVS\cite{murali2023latent}, both of which focus on assessing whether the CVS criteria are properly achieved during laparoscopic cholecystectomy.
As shown in Table~\ref{tab:fine-grained}, SurgWorld outperforms all baselines in both accuracy and F1 score, demonstrating its ability to capture safety-critical procedural cues and the completeness of key operative steps.
Accurate action recognition further plays a central role in improving intraoperative situational awareness, enabling context-specific assistance such as real-time procedural guidance, anticipatory error prevention, and adaptive decision-making.
We then evaluate short-term laparoscopic action recognition on SurgicalActions160 \cite{schoeffmann2018video}. SurgWorld achieves an accuracy of 60.42\% and an F1 score of 59.10\%, consistently outperforms the compared baselines.
This result shows that SurgWorld can distinguish localized and rapidly evolving surgical actions beyond coarse phase-level understanding.
Finally, we evaluate action triplet recognition on CholecT50\cite{nwoye2022rendezvous}, which requires recognizing individual components and systematically modeling their interrelationships to capture complex instrument-tissue interactions. This task is particularly important for real-time surgical assistance, procedural monitoring, and informed intraoperative decision support. SurgWorld achieves state-of-the-art performance across all metrics. For component-level recognition, SurgWorld obtains 92.04\% $AP_{i}$, 60.25\% $AP_{v}$, and 52.28\% $AP_{t}$, outperforming DINOv3 by 5.53\%, 4.00\%, and 6.59\% on instruments, verbs, and targets, respectively. On the most challenging triplet association metric, $AP_{ivt}$, SurgWorld achieves 44.32\%, exceeding SurgMotion by 11.24\%. These results demonstrate that SurgWorld learns more expressive surgical representations that generalize across diverse activity patterns, from isolated action components to structured instrument-verb-target interactions.

\begin{figure}[t]
    \centering
    
    \includegraphics[width=0.75\columnwidth]{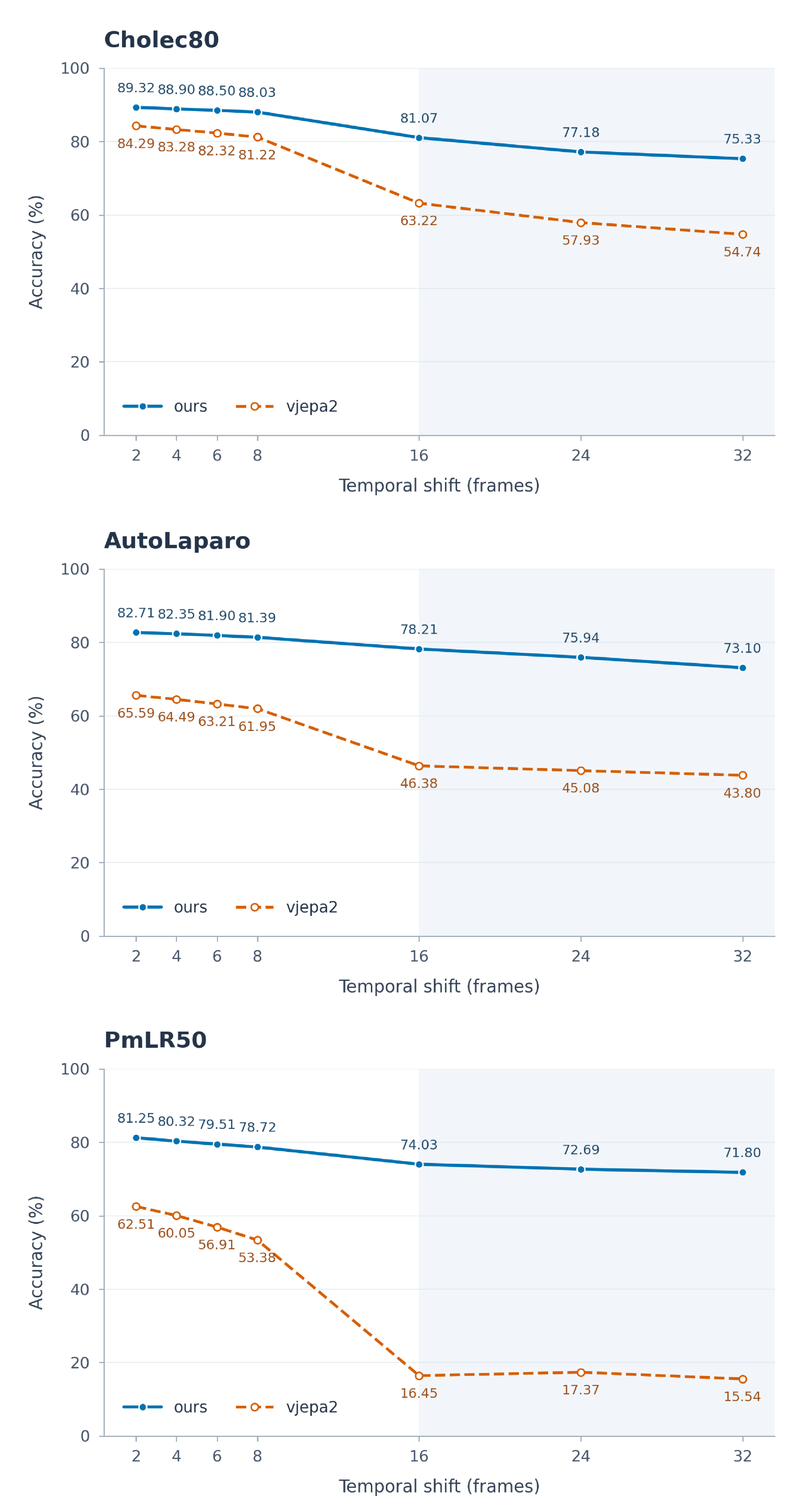}
    \caption{Future latent prediction under varying temporal shifts.}
    \label{fig:fig7}
\end{figure}

\begin{figure}[!t]
    \centering
    \includegraphics[width=0.75\columnwidth]{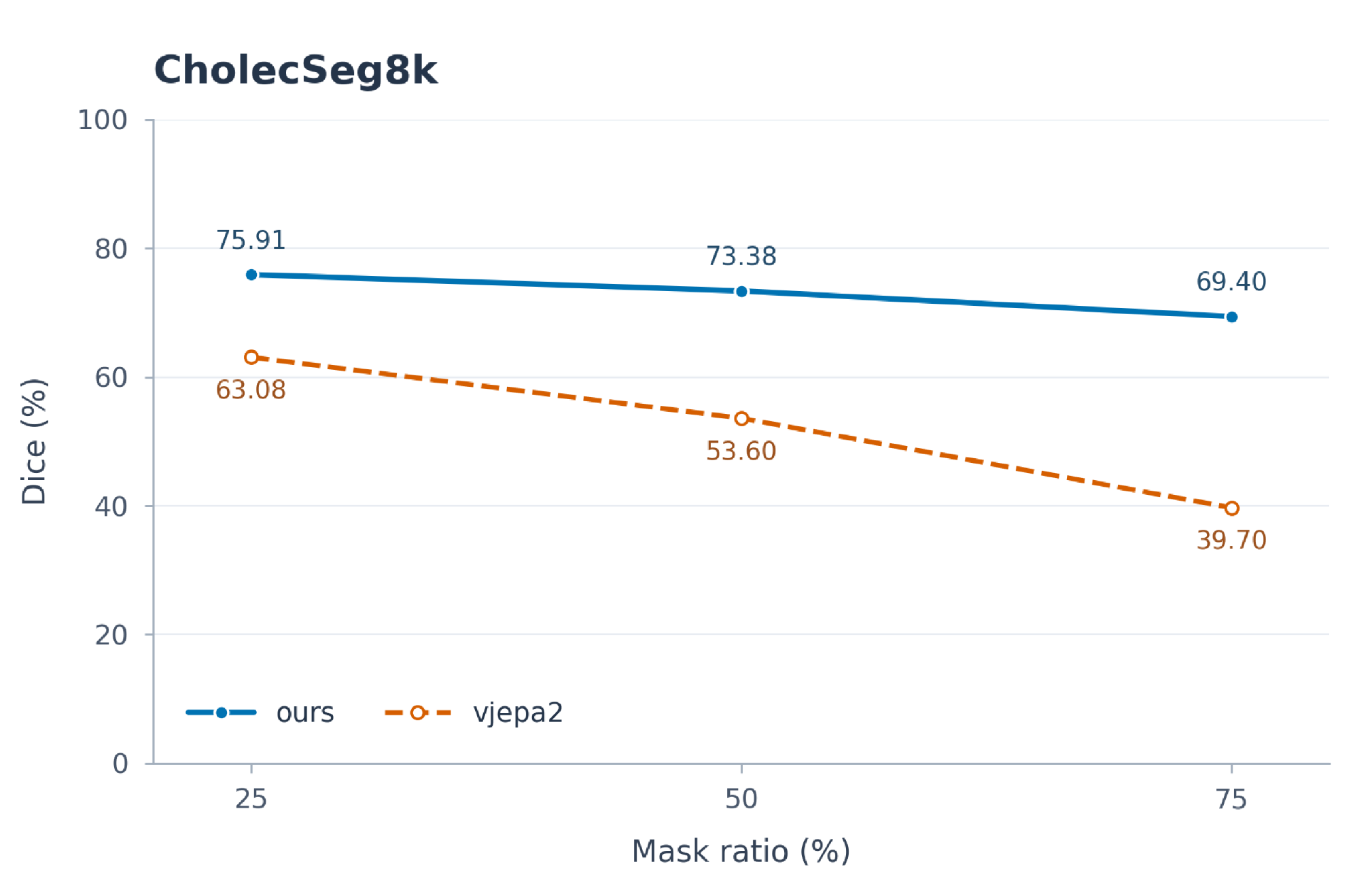}
    
    \caption{
    Dense segmentation under varying spatial masking ratios.
    }
    \label{fig:fig8}
\end{figure}

\begin{table}[t]
    \centering
    \footnotesize
    \setlength{\tabcolsep}{3pt}
    \caption{Results of the pseudo-action sensitivity analysis. Recall values are reported in percentage, with 95\% confidence intervals shown in parentheses.}
    \begin{tabular}{@{}l c c c@{}}
        \toprule
        \textbf{Method}
        & \textbf{Recall@1 (\%)}
        & \textbf{Recall@3 (\%)}
        & \textbf{Recall@5 (\%)} \\
        \midrule

        Random 
        & 1.56
        & 4.68
        & 7.81 \\

        \midrule

        \shortstack[l]{w/o Shift\\Cond.}
        & \shortstack[c]{30.60\\(29.38, 31.84)}
        & \shortstack[c]{52.23\\(50.88, 53.46)}
        & \shortstack[c]{61.77\\(60.66, 62.97)} \\

        \midrule
        
        \textbf{SurgWorld}
        & \shortstack[c]{\textbf{50.63}\\(49.21, 52.02)}
        & \shortstack[c]{\textbf{73.36}\\(71.92, 74.60)}
        & \shortstack[c]{\textbf{82.40}\\(81.22, 83.56)} \\

        \bottomrule
    \end{tabular}
    \label{tab:action-sens}
\end{table}

\begin{figure}[t]
    \centering
    \includegraphics[width=\columnwidth]{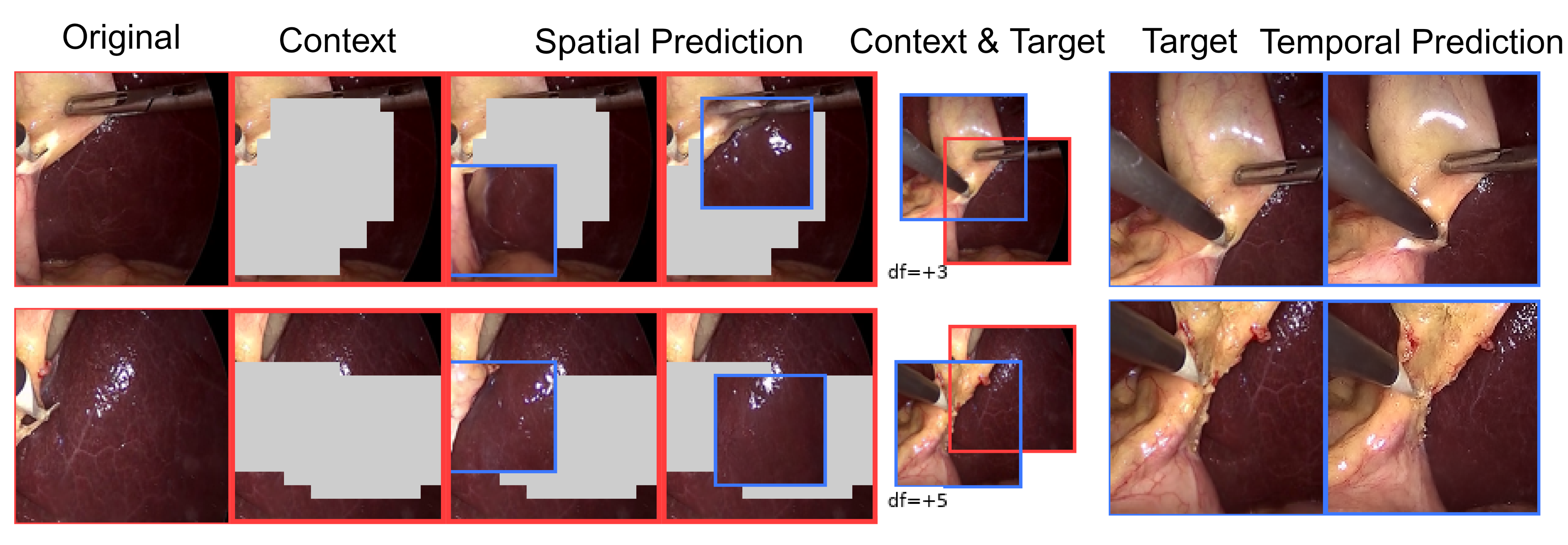}
    
    \caption{
    Visualization of SurgWorld's prediction.
    }
    \label{fig:visualization_surgworld}
\end{figure}

\subsubsection{Surgical dense visual perception}
Surgical AI systems also require dense pixel-level predictions for clinically relevant scene understanding tasks, such as lesion, tissue, and instrument segmentation.
Following established protocols \cite{fan2020pranet}, we evaluate polyp segmentation on two In-distribution (ID) datasets (CVC-ClinicDB \cite{bernal2015wm} and Kvasir \cite{jha2019kvasir}) and three Out-of-distribution (OOD) datasets (CVC-300 \cite{vazquez2017benchmark}, CVC-ColonDB \cite{bernal2012towards}, and ETIS-LaribPolypDB \cite{silva2014toward}), together with surgical semantic segmentation on CholecSeg8k \cite{hong2020cholecseg8k}. As shown in Table~\ref{tab:segmentation}, SurgWorld achieves the highest Dice scores across all six benchmarks. The qualitative results in Fig.~\ref{fig:id_ood_seg} further demonstrate that SurgWorld produces segmentation masks with better object coverage and more accurate boundary delineation compared with competing methods. These results highlight the effectiveness of SurgWorld in learning spatially coherent and transferable representations for dense surgical scene understanding.

\subsubsection{Latent space analysis}
To qualitatively assess the global semantic organization and local spatial structure of the learned representations, we visualize frozen clip-level embeddings using t-SNE following PCA-based dimensionality reduction, and project patch-level dense features into RGB space using PCA. As shown in Fig.~\ref{fig:pca-tsne} (left), SurgWorld exhibits a more structured semantic embedding space, with samples from different surgical procedure categories forming relatively compact and well-separated clusters. Meanwhile, Fig.~\ref{fig:pca-tsne} (right) shows that the PCA projections of dense features exhibit spatially coherent patterns associated with surgical instruments, anatomical structures, and local interaction regions. Together, these visualizations suggest that SurgWorld captures procedure-level semantic distinctions while preserving fine-grained spatial structure in its dense representations.

\begin{table*}[!t]
    \centering
    \footnotesize
    \setlength{\tabcolsep}{2pt}
    
    \caption{Ablation study on the contributions of individual pre-training components.}
    \begin{tabular}{ c c c 
                    | c c 
                    | c c
                    | c c
                    | c c
                    | c c
                    | c c
                    | c c
                    | c
                    | c}
        \toprule

        \multicolumn{3}{c|}{\textbf{Pre-training Components}}
        & \multicolumn{2}{c|}{\textbf{Cataract101}}
        & \multicolumn{2}{c|}{\textbf{Cataract}}
        & \multicolumn{2}{c|}{\textbf{Cataract21}}
        & \multicolumn{2}{c|}{\textbf{M2CAI16}}
        & \multicolumn{2}{c|}{\textbf{AutoLaparo}}
        & \multicolumn{2}{c|}{\textbf{Cholec80}}
        & \multicolumn{2}{c|}{\textbf{PMLR50}}
        & \multicolumn{1}{c}{\textbf{CholecT50}}
        & \multicolumn{1}{c}{\textbf{CholecSeg8k}} 
        \\

        \cmidrule(lr){1-3}
        \cmidrule(lr){4-5}
        \cmidrule(lr){6-7}
        \cmidrule(lr){8-9}
        \cmidrule(lr){10-11}
        \cmidrule(lr){12-13}
        \cmidrule(lr){14-15}
        \cmidrule(lr){16-17}
        \cmidrule(lr){18-18}
        \cmidrule(lr){19-19}

        \textbf{OSDM\_s}
        & \textbf{OSDM\_t}
        & \textbf{GVS}
        & Acc
        & F1
        & Acc
        & F1
        & Acc
        & F1
        & Acc
        & F1
        & Acc
        & F1
        & Acc
        & F1
        & Acc
        & F1
        & AP$_{ivt}$
        & Dice \\

        \midrule

        & 
        &
        & 83.25 & 77.83
        & 70.06 & 64.02
        & 73.21 & 67.71
        & 82.31 & 72.50
        & 77.13 & 66.64
        & 85.98 & 78.78
        & 72.53 & 55.95
        & 35.67
        & 66.65 \\

        \checkmark
        &
        &
        & 90.15 & 86.72
        & 78.71 & 71.31
        & 78.72 & 75.20
        & 82.60 & 73.31
        & 79.63 & 68.16
        & 86.37 & 79.33
        & 70.05 & 53.20
        & 40.29
        & 72.49 \\

        \checkmark
        & \checkmark
        &
        & 91.99 & 88.52
        & 81.39 & 76.43
        & 87.92 & 85.84
        & 84.15 & 75.86
        & 82.63 & 70.12
        & 88.03 & 80.80
        & 76.67 & 61.22
        & 42.53
        & 78.15 \\

        \checkmark
        & \checkmark
        & \checkmark
        & \textbf{92.68} & \textbf{90.98}
        & \textbf{83.02} & \textbf{78.20}
        & \textbf{91.94} & \textbf{90.13}
        & \textbf{84.49} & \textbf{77.06}
        & \textbf{83.04} & \textbf{72.62}
        & \textbf{89.42} & \textbf{81.64}
        & \textbf{82.31} & \textbf{67.68}
        & \textbf{44.32}
        & \textbf{80.36} \\

        \bottomrule
    \end{tabular}
    \label{tab:component_ablation}
\end{table*}

\begin{table*}[t]
    \centering
    \footnotesize
    \setlength{\tabcolsep}{2pt}
    \renewcommand{\arraystretch}{1.0}
    \caption{Ablation study on the effects of temporal shift ranges across the two pre-training stages. Temporal shifts are measured in frames.}
    
    \resizebox{\textwidth}{!}{%
    \begin{tabular}{c | c *{7}{| c c} | c | c}
        \toprule
        \multirow{2}{*}{\shortstack{Pretraining\\stage}}
        & \multirow{2}{*}{\shortstack{Temporal\\shift}}
        & \multicolumn{2}{|c|}{\textbf{Cataract}}
        & \multicolumn{2}{c|}{\textbf{Cataract21}}
        & \multicolumn{2}{c|}{\textbf{M2CAI16}}
        & \multicolumn{2}{c|}{\textbf{AutoLaparo}}
        & \multicolumn{2}{c|}{\textbf{Cataract101}}
        & \multicolumn{2}{c|}{\textbf{Cholec80}}
        & \multicolumn{2}{c|}{\textbf{PMLR50}}
        & \textbf{CholecT50}
        & \textbf{CholecSeg8k} \\
        \cmidrule(lr){3-4}
        \cmidrule(lr){5-6}
        \cmidrule(lr){7-8}
        \cmidrule(lr){9-10}
        \cmidrule(lr){11-12}
        \cmidrule(lr){13-14}
        \cmidrule(lr){15-16}
        \cmidrule(lr){17-17}
        \cmidrule(lr){18-18}
        & & Acc & F1 & Acc & F1 & Acc & F1 & Acc & F1
          & Acc & F1 & Acc & F1 & Acc & F1 & AP$_{\mathrm{ivt}}$ & Dice \\
        \midrule
        \multirow{2}{*}{Stage 1} & 0--8
        & 82.27 & 76.51
        & 90.66 & 89.83
        & 83.91 & 75.71
        & 81.72 & 70.99
        & 90.45 & 89.84
        & 89.24 & 81.41
        & 81.27 & 66.93
        & 42.27
        & 78.77
        \\
         & 0--16
        & 78.15 & 71.98
        & 87.16 & 86.06
        & 82.17 & 72.08
        & 77.96 & 67.46
        & 90.23 & 88.33
        & 85.37 & 78.16
        & 74.09 & 56.55
        & 36.82
        & 76.21
        \\
        \midrule
        \multirow{3}{*}{Stage 2} & 0--8
        & 82.21 & 77.17
        & 90.78 & 88.33
        & 83.22 & 76.38
        & 82.68 & 71.23
        & 91.67 & 89.17
        & 88.61 & 81.51
        & 81.48 & 65.46
        & 42.16
        & 78.23
        \\
         & 0--16
        & \textbf{83.02} & \textbf{78.20}
        & \textbf{91.94} & \textbf{90.13}
        & \textbf{84.49} & \textbf{77.06}
        & \textbf{83.04} & \textbf{72.62}
        & \textbf{92.68} & \textbf{90.98}
        & \textbf{89.42} & \textbf{81.64}
        & \textbf{82.31} & \textbf{67.68}
        & \textbf{44.32}
        & \textbf{80.36}
        \\
         & 0--32
        & 82.05 & 77.09
        & 90.80 & 88.13
        & 83.88 & 74.67
        & 82.36 & 71.96
        & 92.10 & 89.99
        & 87.58 & 80.40
        & 80.61 & 65.14
        & 42.96
        & 77.97
        \\
        \bottomrule
    \end{tabular}%
    }
    \label{tab:temporal_shift_ablation}
\end{table*}

\begin{table*}[t]
    \centering
    \small
    \setlength{\tabcolsep}{2pt}
    \caption{Ablation study on the effect of different GVS teachers}
    \begin{tabular}{
        c
        | c c
        | c c
        | c c
        | c c
        | c c
        | c c
        | c c
        | c
        | c
    }
        \toprule

        \multirow{2}{*}{
            \shortstack{GVS\\Teacher}
        }
        & \multicolumn{2}{c|}{\textbf{Cataract}}
        & \multicolumn{2}{c|}{\textbf{Cataract21}}
        & \multicolumn{2}{c|}{\textbf{M2CAI16}}
        & \multicolumn{2}{c|}{\textbf{AutoLaparo}}
        & \multicolumn{2}{c|}{\textbf{Cataract101}}
        & \multicolumn{2}{c|}{\textbf{Cholec80}}
        & \multicolumn{2}{c|}{\textbf{PMLR50}}
        & \multicolumn{1}{c|}{\textbf{CholecT50}}
        & \multicolumn{1}{c}{\textbf{CholecSeg8k}} \\

        \cmidrule(lr){2-3}
        \cmidrule(lr){4-5}
        \cmidrule(lr){6-7}
        \cmidrule(lr){8-9}
        \cmidrule(lr){10-11}
        \cmidrule(lr){12-13}
        \cmidrule(lr){14-15}
        \cmidrule(lr){16-16}
        \cmidrule(lr){17-17}

        &
        Acc & F1
        & Acc & F1
        & Acc & F1
        & Acc & F1
        & Acc & F1
        & Acc & F1
        & Acc & F1
        & AP$_{ivt}$
        & Dice \\

        \midrule

        VideoMAE
        & 82.48 & 77.58
        & 91.93 & \textbf{90.77}
        & \textbf{86.03} & \textbf{77.08}
        & 82.68 & 70.96
        & \textbf{92.94} & \textbf{91.06}
        & 89.01 & \textbf{81.79}
        & \textbf{84.04} & \textbf{68.08}
        & 42.72
        & 78.83 \\

        DINOv3
        & \textbf{83.02} & \textbf{78.20}
        & \textbf{91.94} & 90.13
        & 84.49 & 77.06
        & \textbf{83.04} & \textbf{72.62}
        & 92.68 & 90.98
        & \textbf{89.42} & 81.64
        & 82.31 & 67.68
        & \textbf{44.32}
        & \textbf{80.36} \\

        \bottomrule
    \end{tabular}
    \label{tab:dense_teacher}
\end{table*}

\begin{table*}[t]
    \centering
    \footnotesize
    \setlength{\tabcolsep}{2pt}
    \renewcommand{\arraystretch}{1.0}
    \caption{Ablation study on GVS supervision configurations, including additional supervision on cross-view prediction branches and sensitivity to the GVS loss weight}
    
    \resizebox{\textwidth}{!}{%
    \begin{tabular}{c | c *{7}{| c c} | c | c}
        \toprule
        \multirow{2}{*}{\shortstack{Extra\\GVS}}
        & \multirow{2}{*}{\shortstack{GVS\\weight}}
        & \multicolumn{2}{|c|}{\textbf{Cataract}}
        & \multicolumn{2}{c|}{\textbf{Cataract21}}
        & \multicolumn{2}{c|}{\textbf{M2CAI16}}
        & \multicolumn{2}{c|}{\textbf{AutoLaparo}}
        & \multicolumn{2}{c|}{\textbf{Cataract101}}
        & \multicolumn{2}{c|}{\textbf{Cholec80}}
        & \multicolumn{2}{c|}{\textbf{PMLR50}}
        & \textbf{CholecT50}
        & \textbf{CholecSeg8k} \\
        \cmidrule(lr){3-4}
        \cmidrule(lr){5-6}
        \cmidrule(lr){7-8}
        \cmidrule(lr){9-10}
        \cmidrule(lr){11-12}
        \cmidrule(lr){13-14}
        \cmidrule(lr){15-16}
        \cmidrule(lr){17-17}
        \cmidrule(lr){18-18}
        & & Acc & F1 & Acc & F1 & Acc & F1 & Acc & F1
          & Acc & F1 & Acc & F1 & Acc & F1 & AP$_{\mathrm{ivt}}$ & Dice \\
        \midrule
        No & 2.0
        & 78.87 & 74.56
        & 90.63 & 88.91
        & 83.77 & 74.32
        & 80.97 & 71.25
        & 92.64 & 90.56
        & 88.08 & 81.06
        & 73.40 & 57.38
        & 41.06
        & 77.19
        \\
        \midrule
        \multirow{5}{*}{Yes} 
        & 0.2
        & 79.88 & 74.61
        & 90.18 & 88.53
        & 83.64 & 75.52
        & 80.61 & 69.76
        & 91.91 & 89.51
        & 87.75 & 80.36
        & 77.67 & 59.47
        & 40.54
        & 77.70
        \\
        & 1.0
        & 81.22 & 76.37
        & 91.16 & 89.77
        & 85.32 & 75.94
        & 82.31 & \textbf{73.63}
        & 93.63 & 91.97
        & 89.44 & \textbf{82.52}
        & 79.45 & 63.66
        & 40.27
        & 81.51
        \\
        & 2.0
        & \textbf{83.02} & \textbf{78.20}
        & \textbf{91.94} & \textbf{90.13}
        & 84.49 & \textbf{77.06}
        & 83.04 & 72.62
        & 92.68 & 90.98
        & 89.42 & 81.64
        & \textbf{82.31} & \textbf{67.68}
        & \textbf{44.32}
        & 80.36
        \\
        & 5.0
        & 81.20 & 75.28
        & 90.78 & 88.90
        & \textbf{86.07} & 76.76
        & \textbf{83.39} & 73.25
        & 93.38 & 91.28
        & 89.56 & 82.31
        & 78.73 & 64.76
        & 40.36
        & 80.59
        \\
        & 10.0
        & 80.39 & 75.70
        & 91.40 & 89.11
        & 82.14 & 75.19
        & 82.42 & 71.90
        & \textbf{93.93} & \textbf{92.27}
        & \textbf{89.52} & 82.39
        & 81.43 & 66.42
        & 40.14
        & \textbf{82.15}
        \\
        \bottomrule
    \end{tabular}%
    }
    \label{tab:ab4}
\end{table*}

\subsubsection{Data efficiency}
We further assess the data efficiency of SurgWorld by fine-tuning downstream models with varying proportions of labeled training data (5\%, 10\%, 20\%, 50\%, and 100\%) on Cholec80. As shown in Fig.~\ref{fig:result}, SurgWorld consistently outperforms the competing methods across different data regimes. Notably, with only 10\% of the labeled training data, SurgWorld achieves higher accuracy and a comparable F1 score to those of competing methods trained on the full dataset, demonstrating its strong label efficiency. 

To further examine the structure of the learned representations, we visualize the feature embeddings using Uniform Manifold Approximation and Projection (UMAP) in Fig.~\ref{fig:umap}. With only 10\% of the labeled training data, SurgWorld exhibits more distinct clusters corresponding to the seven surgical phases than the competing methods. This qualitative observation complements the quantitative results, suggesting that SurgWorld learns representations that facilitate surgical phase discrimination under limited supervision and may help reduce the annotation requirements for surgical video understanding.

\subsection{Evaluating Latent Prediction Beyond Observation}

\subsubsection{Future state prediction under partial observations}

To evaluate whether SurgWorld can predict task-relevant representations beyond directly observed inputs, we conduct the experiments illustrated in Fig.~\ref{fig:fig1}(b). For temporal understanding, we first sample a 16-frame target clip $y$ and obtain its preceding 16-frame context clip $x$ by shifting the temporal sampling window backward by $\Delta t$ frames relative to $y$. Conditioned on $x$ and $\Delta t$, SurgWorld predicts the latent representation of the entire target clip, which is subsequently used for downstream classification. We evaluate temporal shifts of $\Delta t \in \{2,4,6,8,16,24,32\}$, where the context and target clips become fully non-overlapping when $\Delta t \geq 16$. As shown in Fig.~\ref{fig:fig7}, SurgWorld consistently outperforms the matched V-JEPA2 baseline across all evaluated temporal shifts on all three benchmarks. Notably, at $\Delta t=8$, where half of the target clip lies beyond the observed context, SurgWorld surpasses the strongest baseline evaluated using fully observed inputs, as shown by comparing with Table~\ref{tab:workflow}. For dense visual perception, $r\%$ of the input regions are randomly masked. Conditioned on the relative spatial displacement $(\Delta x,\Delta y)$, SurgWorld predicts the latent representations of the masked regions and restores them to their original locations, yielding a complete feature map for evaluation with a frozen segmentation head. As shown in Fig.~\ref{fig:fig8}, SurgWorld consistently outperforms the matched V-JEPA2 baseline across all evaluated masking ratios. Notably, even with 50\% of the input regions masked, SurgWorld achieves segmentation performance comparable to that of the strongest surgical baseline evaluated using fully observed inputs, as shown by comparing with Table~\ref{tab:segmentation}. These results demonstrate that SurgWorld can effectively predict task-relevant representations beyond directly observed inputs across both temporal and spatial dimensions, enabling robust downstream understanding under partial observations.

\subsubsection{Pseudo-action sensitivity analysis}

We design a pseudo-action sensitivity test to evaluate whether SurgWorld can use the given action to identify the corresponding out-of-sight surgical state.
Specifically, given a context–target pair $(x,y)$ from the same video, we define their relative spatio-temporal shift as $a$, forming a triplet $(x,y,a)$.
For each target, we sample $N$ distractors from other spatio-temporal locations.
Conditioned on $x$ and $a$, SurgWorld predicts the target latent representation, and all candidates are ranked by their feature distance to the prediction. We report the top-$k$ recall of the ground-truth target. 
As shown in Table~\ref{tab:action-sens}, SurgWorld achieves an average top-$5$ recall of 82.40, over $10\times$ random choice, while removing the shift condition $a$ substantially degrades performance, indicating the importance of explicit action conditioning to SurgWorld’s predictive capability.

\subsubsection{Visualization of predicted latent states}
We train a generative decoder that maps average-pooled predictor outputs to pixel space following \cite{assran2023self}. As shown in Fig.~\ref{fig:visualization_surgworld}, the reconstructed states preserve anatomical structures, surgical instruments, and fine-grained interactions, demonstrating globally coherent and locally detailed prediction of future surgical states.

\subsection{Ablation Study}
We conduct four ablation studies to systematically investigate the effectiveness of the proposed components and key design choices in SurgWorld: (1) the contribution of each pre-training component, including spatial OSDM, temporal OSDM, and GVS; (2) the effect of different temporal shift ranges across the two pre-training stages; (3) the selection of the dense-feature teacher for GVS; (4) the effects of GVS supervision configurations, including additional supervision on cross-view prediction branches and the choice of loss weight.

\subsubsection{Component Analysis} To assess the incremental contributions of the proposed components, we conduct a progressive ablation study, starting from a V-JEPA2 baseline retrained on the same surgical video corpus and sequentially incorporating spatial OSDM, temporal OSDM, and GVS. As shown in Table~\ref{tab:component_ablation}, incorporating spatial OSDM yields substantial improvements in downstream performance, demonstrating the effectiveness of spatially conditioned latent prediction for surgical representation learning. Building upon spatial OSDM, the addition of temporal OSDM brings further improvements across surgical workflow recognition, fine-grained action understanding, and dense visual perception, highlighting the complementary benefits of jointly modeling spatial and temporal variations. Finally, GVS introduces complementary geometric supervision, yielding additional performance gains, particularly in dense visual perception. Overall, these results demonstrate the effectiveness of spatio-temporal OSDM in learning transferable surgical representations and the complementary role of GVS in enhancing spatial coherence and fine-grained visual understanding.

\subsubsection{Effect of Temporal Shift Ranges Across Pre-training Stages} We investigate the influence of temporal shift ranges on representation learning across the two pre-training stages. As shown in Table~\ref{tab:temporal_shift_ablation}, during Stage 1, which operates on 16-frame clips, increasing the maximum temporal shift from 8 to 16 frames consistently degrades downstream performance. This suggests that excessively large temporal shifts during early pre-training may weaken the correspondence between context and target states, making latent prediction more challenging. During Stage 2, which employs longer 64-frame clips, we further investigate maximum temporal shifts of 8, 16, and 32 frames while keeping the Stage 1 configuration fixed. A maximum shift of 16 frames achieves the best performance across all evaluated benchmarks, whereas further increasing it to 32 frames leads to noticeable degradation. These findings suggest that the optimal temporal shift range depends on the training stage, with relatively small shifts facilitating initial representation learning and moderately larger shifts enabling the model to capture richer temporal variations while maintaining reliable context–target correspondence. Furthermore, compared with the Stage 1 checkpoint, the final model consistently improves performance across all evaluated benchmarks, demonstrating the effectiveness of the subsequent long-clip cooldown training.

\subsubsection{The selection of the dense-feature teacher for GVS}  We investigate the influence of different dense-feature teachers on SurgWorld by comparing VideoMAE and DINOv3 under the same pre-training configuration. As shown in Table~\ref{tab:dense_teacher}, both teachers yield comparable overall performance across downstream benchmarks. We therefore adopt DINOv3 as the dense-feature teacher for GVS. Notably, as shown in Tables~\ref{tab:workflow}, \ref{tab:fine-grained} and \ref{tab:segmentation}, SurgWorld consistently outperforms the standalone DINOv3 and VideoMAE baselines across all evaluated downstream metrics, demonstrating the effectiveness of integrating geometric supervision with spatio-temporal predictive learning.

\subsubsection{Effect of GVS Supervision Configurations} We investigate two key design choices of GVS: the application of additional geometric supervision to cross-view prediction branches and the selection of the GVS loss weight. As shown in Table~\ref{tab:ab4}, extending GVS supervision to cross-view predictions consistently improves performance across all evaluated benchmarks. We further investigate the sensitivity of SurgWorld to the GVS loss weight by evaluating five values: $\lambda_{\mathrm{GVS}} \in \{0.2, 1.0, 2.0, 5.0, 10.0\}$. Considering the overall performance across different downstream benchmarks, we adopt $\lambda_{\mathrm{GVS}} = 2.0$ as the default setting for SurgWorld.

\section{Conclusion}

In this paper, we introduce SurgWorld, a video-native, self-supervised framework for surgical representation learning that extends beyond directly observed content through spatio-temporal shift-conditioned latent prediction.  By jointly integrating OSDM and GVS objectives within an agglomerative training framework, SurgWorld learns unified representations of surgical dynamics and visual geometry. Despite being pretrained exclusively on publicly available surgical videos, it effectively predicts out-of-sight surgical states and achieves strong performance across 17 surgical benchmarks. These results advance surgical video pretraining beyond visible observation encoding toward predictive modeling of partially observable surgical environments.

\section*{REFERENCES}
\bibliographystyle{IEEEtran} 
\bibliography{IEEEtran}

\end{document}